\pdfoutput=1

\documentclass[11pt]{article}

\usepackage[]{EACL2023}

\usepackage{times}
\usepackage{latexsym}

\usepackage[T1]{fontenc}

\usepackage[utf8]{inputenc}

\usepackage{microtype}

\usepackage{inconsolata}

\usepackage{url}
\usepackage{booktabs}
\usepackage{caption}
\usepackage{subcaption}
\usepackage{verbatim}
\usepackage[export]{adjustbox}

\newcommand{\ignore}[1]{}

\newcommand{\ie}{\emph{i.e.}}
\newcommand{\eg}{\emph{e.g.}}

\newcommand{\etc}{\emph{etc.}}

\usepackage{amsmath}
\usepackage{cleveref}
\usepackage{refstyle}
\crefname{section}{Sec.}{}
\crefname{table}{Tab.}{}
\crefname{figure}{Fig.}{}
\crefname{algorithm}{Algorithm}{}
\crefname{equation}{Equation}{}
\crefname{appendix}{App.}{}
\crefname{prop}{Proposition}{}
\crefname{thm}{Theorem}{}

\usepackage{multirow}
\usepackage{colortbl}
\newcommand{\band}{\rowcolor{gray!10}}
\usepackage{enumitem}

\usepackage{xspace}

\usepackage{euflag}

\newcommand{\vl}{V\&L}
\newcommand{\albef}{\textsc{ALBEF}\xspace}
\newcommand{\dalbef}{\textsc{ALBEF}$_\text{DISC}$\xspace}
\newcommand{\galbef}{\textsc{ALBEF}$_\text{GEN}$\xspace}

\newcommand{\vilbert}{\textsc{ViLBERT}\xspace}
\newcommand{\dvilbert}{\textsc{ViLBERT}$_\text{DISC}$\xspace}
\newcommand{\gvilbert}{\textsc{ViLBERT}$_\text{GEN}$\xspace}
\newcommand{\topk}{top-$k$}
\newcommand{\vqavii}{{\sc VQAv2}\xspace}
\newcommand{\vizwiz}{{\sc VizWiz}\xspace}
\newcommand{\visualgenome}{{\sc Visual Genome}\xspace}
\newcommand{\vg}{{\sc VG}\xspace}
\newcommand{\gqa}{{\sc GQA}\xspace}
\newcommand{\vqacp}{{\sc VQA-CP}\xspace}

\title{Reassessing Evaluation Practices in Visual Question Answering:\\A Case Study on Out-of-Distribution Generalization}

\newcommand{\ku}{\triangle}
\newcommand{\dm}{\diamondsuit}

\newcommand{\ox}{\natural}

\author{Aishwarya Agrawal\Thanks{ denotes equal first author contribution. $^\dagger$ denotes equal contribution. $^\ddagger$ denotes equal senior contribution. Detailed contributions follow at the end of the paper.}$^{~,\ddagger,\dm}$~ Ivana Kaji\'{c}$^{*,\dm}$~ Emanuele Bugliarello$^{*,\ku}$\\
\textbf{Elnaz Davoodi$^{\dagger,\dm}$~ Anita Gergely$^{\dagger,\dm}$~ Phil Blunsom$^{\ox}$~ Aida Nematzadeh$^{*,\ddagger,\dm}$}\\[4pt]
        $^{\dm}$DeepMind~\;$^{\ku}$University of Copenhagen~\;~$^{\ox}$University of Oxford\\[4pt]
        \texttt{\{aiagrawal,kivana,nematzadeh\}@deepmind.com}~\;~
        \texttt{emanuele@di.ku.dk}
       }

\begin{document}
\maketitle

\begin{abstract}
Vision-and-language (V\&L) models pretrained on large-scale multimodal data have demonstrated strong performance on various tasks such as image captioning and visual question answering (VQA). The quality of such models is commonly assessed by measuring their performance on unseen data that typically comes from the same distribution as the training data. However, when evaluated under out-of-distribution (out-of-dataset) settings for VQA, we observe that these models exhibit poor generalization. We comprehensively evaluate two pretrained V\&L models under different settings (i.e. classification and open-ended text generation) by conducting cross-dataset evaluations. We find that these models tend to learn to solve the benchmark, rather than learning the high-level skills required by the VQA task. We also find that in most cases generative models are less susceptible to shifts in data distribution compared to discriminative ones, and that multimodal pretraining is generally helpful for OOD generalization. Finally, we revisit assumptions underlying the use of automatic VQA evaluation metrics, and empirically show that their stringent nature repeatedly penalizes models for correct responses. 
\end{abstract}

\section{Introduction}
\label{sec:intro}

Visual Question Answering (VQA) is the task of automatically answering natural language open-ended questions about images. Tackling VQA involves multiple skills, such as language and visual understanding, integrating information between the two (vision and language) modalities, and commonsense and knowledge based reasoning. One of the goals of the VQA research has been fostering the development of systems that are able to answer \emph{any open-ended question about any image}. This motivation has inspired a fruitful line of research in designing VQA benchmarks \cite[\eg,][]{NIPS2014_d516b136,antol2015vqa,krishna2017visual,balanced_vqa_v2, Hudson_2019_CVPR} and models \cite[\eg,][]{SAN, anderson2018bottom, lu2019vilbert, VL-T5}.

In this work, we investigate if recent pretrained VQA models can indeed answer any open-ended question about images or if they are mostly suitable for answering questions from the VQA benchmarks they are optimized for. 
In other words, \emph{are models learning to solve the task or learning to solve the datasets?} We believe the former is more aligned with the goal of building real-world VQA systems.

To measure whether models learn to solve the task of VQA, we believe we need to examine their \emph{out-of-distribution (OOD)} generalization capabilities: how they perform on examples drawn from a distribution other than that of the training set. 
In this work, we extensively evaluate OOD generalization of current pretrained \vl{} models by conducting cross-dataset evaluations (without any adaptation to the test domain). 

Through our extensive experiments, we provide in-depth discussion on the following questions:
\begin{itemize}[noitemsep, topsep=1pt]
    \item \emph{How well do recent models generalize under OOD settings?} We observe a notable drop in performance from IID to OOD settings across models and benchmarks, demonstrating that models mostly learn to solve specific benchmarks as opposed to learning general skills for answering questions about images. This result is not simply due to a mismatch between the set of answers between the training and test VQA datasets, nor due to poor representation of test answers in VQA training data.
    \item \emph{Is multimodal pretraining beneficial for OOD generalization?} We find that while image--text pretraining is helpful in most OOD settings, it is not always more useful than in IID ones. Moreover, it is least useful for OOD evaluation on the \vizwiz benchmark, highlighting the challenges of a real-world benchmark. 
    \item \emph{Is generative modeling more robust to distribution shifts?} In most cases, we observe that generative models---which are not bound to predictions over a fixed set of answers curated from the training data---are more robust to OOD evaluation than discriminative (\ie, classification-based) ones. Moreover, we quantify what the limitations of discriminative models are for real-world VQA applications (\eg, answering questions of visually-impaired users), where the answers a deployed model needs to produce cannot be pre-determined. 
    \item \emph{Are current automatic VQA metrics too stringent for OOD evaluation?} We examine if the performance of our pretrained models is negatively impacted by the current standard VQA accuracy metrics, which match predicted answer strings to a limited number of ground-truth answers. Human evaluation reveals the stringent nature of such accuracy metrics, which is especially pronounced in the OOD settings. Nevertheless, while the IID-to-OOD performance gap is reduced after human evaluation, models still exhibit poor generalization to OOD VQA benchmarks.
\end{itemize}
We believe our OOD evaluations and supporting analyses expose the shortcomings of current models, and recommend future work to adopt these evaluation practices to provide real-world, robust assessment of VQA systems.

\section{Related Work}
\label{sec:related_work}

\textbf{Beyond IID evaluation in VQA.} 
Previous work has evaluated VQA models beyond the IID setting for robustness to \emph{specific and controlled} aspects -- novel compositions of seen concepts \cite{Agrawal2017CVQAAC, Johnson_2017_CVPR, Hudson_2019_CVPR}, change in prior distributions of answers per question type \cite{Agrawal_2018_CVPR, gokhale2020mutant, niu2021counterfactual}, adversarial examples provided by humans \cite{fb_adv_vqa, msr_adv_vqa}, consistency, negation, and simple perturbation in questions \cite{Jimenez2022CARETSAC}, counter-examples \cite{Dancette2021BeyondQB}, and controlled shifts in language and vision modalities \cite{akula-etal-2021-crossvqa}. 
Our focus, however, is to evaluate for \emph{overall} robustness to OOD data without controlling for specific aspects, by testing our models on different OOD benchmarks. We believe our experimental setting more closely emulates the expected experience of deployed VQA systems. Moreover, when the exact nature of distribution shift between train and test splits is known (such as in \cite{Agrawal_2018_CVPR}), approaches developed to tackle such shifts tend to rely on the explicit knowledge of construction of such OOD splits resulting in inflated sense of progress \cite{teney2020value}.

Similar to us, \citet{Zhang2021DomainrobustVW, Hudson_2019_CVPR} also present some experimental results on VQA OOD evaluation, however they do it in limited manner (\eg, do not consider all pairs of datasets, do not evaluate the effect of multimodal pretraining, etc.). To our best knowledge, ours is the first work to extensively quantifying the extent of IID to OOD performance drops in current VQA models and study the effect of several factors: answer overlap, multimodal pretraining, generative vs. discriminative modeling, and stringent evaluation metric.

\textbf{Domain adaptation in VQA.}
Some studies \cite{Jabri_2016_ECCV, Chao_2018_CVPR, Zhang2021DomainrobustVW} have explored domain adaptation of VQA models from one VQA benchmark to another. 
Our focus, instead, is on evaluating \emph{zero-shot} cross-benchmark generalization \emph{without} any adaptation.
This allows us to assess the robustness of current models towards unforeseen distribution shifts. 
Our work is similar to that of \citet{name_the_dataset} and \citet{hendrycks-etal-2020-pretrained}, who study OOD generalization in vision and text.

\textbf{Zero-shot VQA with pretrained models.}
In an emerging line of research \cite{frozen, alayrac2022flamingo, song2022clip, piergiovanni2022answer}, large-scale pretrained unimodal models \cite{GPT-3,CLIP} are repurposed to tackle VQA in zero-shot or few-shot fashion. 
While such zero-shot VQA evaluations are a better test of generalization than IID evaluations, our focus, differently, is on investigating whether models can generalize to unseen datasets \emph{upon being taught the task by showing examples from one dataset}. Moreover, this line of work does not focus on a thorough analysis of models in OOD settings (which is hard to define for these models due to the massive amount of data they are pretrained on).

\section{Experimental Setup}
\label{sec:exp_setup}

In this section, we present our framework to examine OOD generalization in VQA.
We examine two pretrained Transformers across five benchmarks.

\subsection{Models} 
We evaluate the performance of two representative, widely-used pretrained models that have achieved strong performance in various \vl{} tasks in the last few years: \vilbert~\cite{lu2019vilbert} and \albef~\cite{li2021align}.
We evaluate these models in a broad range of settings (generative/discriminative, w/wo pretraining, and multiple benchmarks), resulting in 128 experiments.
We chose these models as they include components shown to be important in the literature: cross-attention (\vilbert{} and \albef), and contrastive learning (\albef).
We note that our goal is to study trends that hold across different models, and we leave for future work controlled comparisons across architectures.

\vspace{1.7mm}
\noindent
\textbf{\vilbert} is one of the first, yet strong models in the recent pretrain--fine-tune paradigm in \vl.
Its inputs are a sequence of sub-word tokens~\cite{wu2016google}, and a set of regions of interest given by a Faster R-CNN~\cite{ren2015faster,anderson2018bottom}.
The authors fine-tune it on \vqavii by learning a classifier over the most frequent answers.
We first re-implement this model successfully, and then extend it to a generative setting by pretraining and fine-tuning a Transformer decoder (more details in \cref{app:exp_details}).
We denote the discriminative/generative version as \dvilbert{}/\gvilbert.
Unless otherwise specified, results for \dvilbert{} are from our code base for direct comparison with \gvilbert.

\textbf{\albef} is a state-of-the-art \vl{} encoder whose  visual inputs are image patches encoded by a vision Transformer~\cite{dosovitskiy2021an,touvron2021training} that is jointly trained with the rest of the model.
\citet{li2021align} fine-tune \albef on \vqavii by adding a 6-layer Transformer decoder to generate answers (\galbef).
We use the official implementation,\footnote{\url{https://github.com/salesforce/ALBEF}.} and furthermore train a discriminative variant (\dalbef) by learning a multi-answer classifier, as in \dvilbert.

\vspace{1.7mm}
\noindent
In our analysis, we investigate the role of multimodal pretraining.
\vilbert{} was pretrained on 3M image--text pairs from Conceptual Captions (CC;~\citealt{sharma-etal-2018-conceptual}).
\citet{li2021align} released two checkpoints for \albef: one pretrained on 4M images from CC, MS-COCO~\cite{lin2014microsoft}, SBU~\cite{ordonez2011im2text} and Visual Genome~\cite{krishna2017visual}; the other one is further pretrained on Conceptual 12M~\cite{changpinyo2021conceptual} for a total of 14M images.\footnote{We also conducted experiments with \vilbert{} pretrained on same datasets as the 4M \albef checkpoint. We found no significant difference compared to the results presented throughout this paper.}

\subsection{Datasets and Evaluation Metrics} \label{subsec:dataset_eval_metric}

\begin{table}[t]
     \setlength{\tabcolsep}{2.5pt}
     \centering
     \small
     \resizebox{0.48\textwidth}{!}{
         \begin{tabular}{@{}l@{\hspace{4\tabcolsep}}rr|rr}
             \toprule
             \textbf{Dataset} & \textbf{\# Train (imgs / qns)} & \textbf{\# Val (imgs / qns)} & \textbf{\# Classes} & \textbf{Coverage [\%]}\\
             \midrule   
             \vqavii & 82{,}783 / 443{,}757 & 40{,}504 / 214{,}354 & 3{,}129 & 98.07 / 98.07 \\
             \gqa    & 72{,}140 / 943{,}000 & 10{,}234 / 132{,}062 & 1{,}533 & 99.78 / 99.79 \\
             \vg     & 59{,}635 / 868{,}259 & 39{,}645 / 577{,}063 & 3{,}449 & 76.55 / 76.55 \\
             \vizwiz & 20{,}523 / ~~20{,}523 & 4{,}319 / ~~~~4{,}319 & 3{,}112 & 96.76 / 97.01 \\
             \bottomrule
         \end{tabular}
     }
     \vspace{-0.2cm}
     \caption{Datasets statistics. \#classes is the number of classes we use for the discriminative models; coverage is the percentage of questions that can be answered with our selected classes in train/validation splits.} \label{tab:dataset_stats}
     \vspace{-0.4cm}
\end{table}

\begin{figure*}[t!]
\centering
    \includegraphics[width=\textwidth]{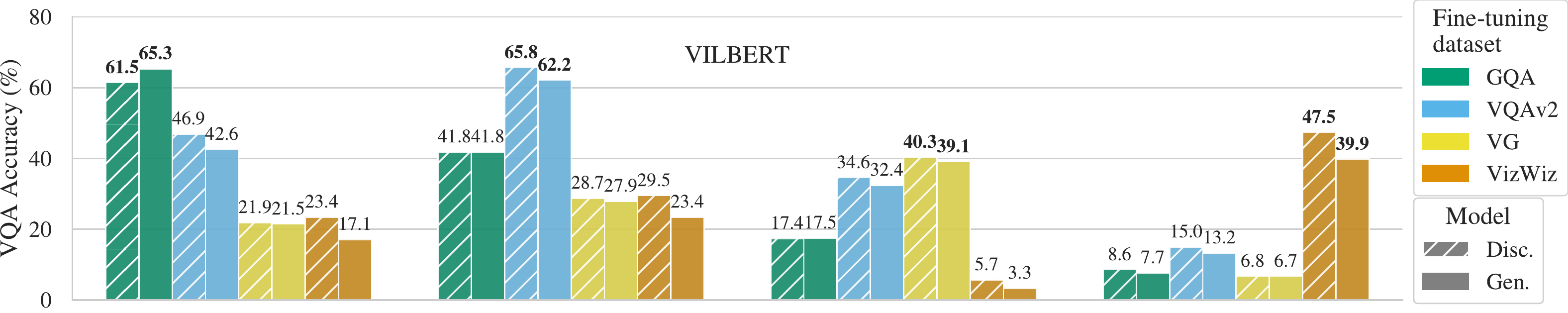}
    \includegraphics[width=0.9\textwidth, left]{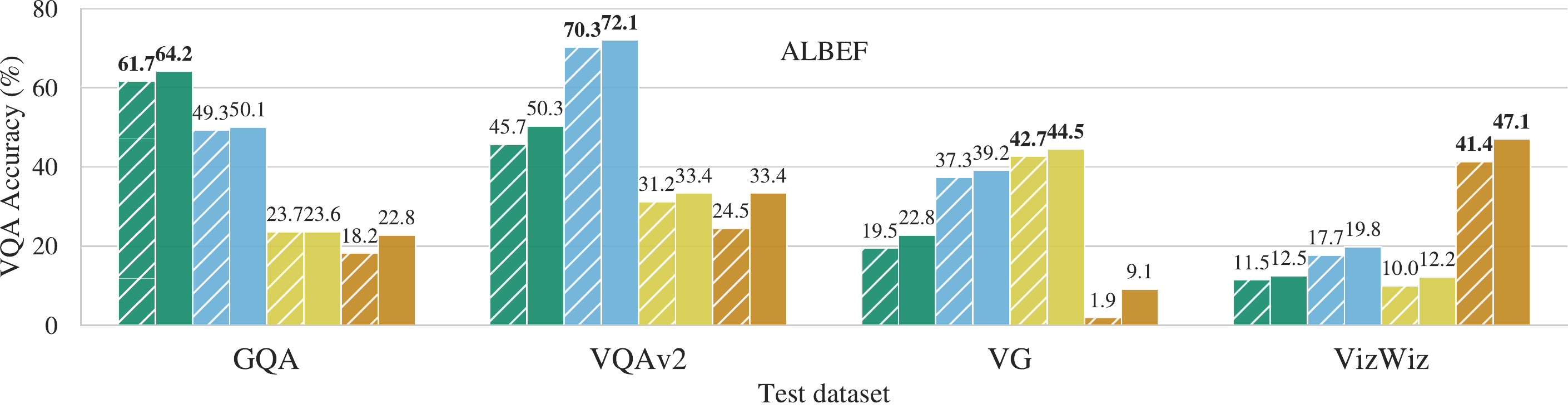}%
\vspace{-0.25cm}
\caption{
IID (highlighted in bold) vs. OOD performance. Top: \vilbert pretrained on CC. Bottom: \albef pretrained on CC, VG, SBU, MS-COCO and C12M datasets. All models are initialized with BERT weights.}
\label{fig:iid_ood}
\vspace{-0.3cm}
\end{figure*}

\paragraph{Datasets.}
We ground our analysis on five diverse VQA datasets: \vqavii~\cite{balanced_vqa_v2}, \gqa~\cite{Hudson_2019_CVPR}, \visualgenome~(\vg;~\citealt{krishna2017visual}), \vizwiz~\cite{Gurari_2018_CVPR} and \vqacp~\cite{Agrawal_2018_CVPR}.
\vqavii is the most commonly used VQA dataset to date.
\vqacp re-splits it such that, for every question type, train and test sets have different prior distributions of answers.
\vg includes questions centered around either the full image or a specific region.
\gqa is a large-scale dataset that focuses on compositionality of template-generated questions.
Finally, \vizwiz is the only real-world VQA dataset, collected from visually-impaired people.
\vg and \gqa have one answer per question, while the other datasets include 10 answers per question.
See \cref{tab:dataset_stats} and \cref{app:exp_details} for more details.

There are several differences among these datasets.
Both \vqavii and \gqa mostly have one-word answers (89\% and 81\%, respectively) whilst there are fewer in \vg (57\%) and \vizwiz (67\%).
The type of questions also varies: \vg does not contain binary `yes/no' questions, but rather spans 6 WH-questions.
By design, \gqa questions require more compositional skills but do not test for counting; while \vizwiz questions are more conversational as they were collected through a speech interface and has a significant proportion of OCR questions (21\%).
Moreover, a significant number of \vizwiz questions (28\%) are \emph{unanswerable} due to the challenges faced by the visually-impaired users in taking pictures, resulting in poor focus, poor lighting or entirely missing the entity of interest. 
As such, the distribution of images in \vizwiz is different from other datasets.

\paragraph{Evaluation metrics.}
These VQA benchmarks compute model accuracy between its prediction and the ground-truth answer(s) by string matching (after simple pre-processing).
\vqavii and \vizwiz, each with 10 answers per question, account for diversity in ground-truth answers by scoring a given model answer as $\texttt{min}\{1.0, 0.3\times\texttt{count}\}$, where \texttt{count} is the number of annotators that used that answer.
For \gqa and \vg, both with one answer per question, we use top-1 accuracy.\footnote{We note that \gqa and \vg propose top-5 accuracy. We, instead, opt for top-1 accuracy to keep a consistent setup with \vqavii and \vizwiz. And we believe top-5 accuracy is impractical for many applications, such as answering questions for visually-impaired users.}

\subsection{Training Details} 

Following common practice, for discriminative models, we select the top-$k$ most frequent answers as the set of answer classes to perform classification over. 
Here $k$ is a dataset-dependent variable, chosen to cover most of the questions (see \cref{tab:dataset_stats}).
All models are trained on the respective training sets and evaluated on the validation sets.
For \vg, we randomly split the data into training and validation (60\%/40\%) with no image is in both splits.
\section{Out-of-Distribution Generalization}
\label{sec:ood_gen}

We examine to what extent our models learn to solve a specific VQA benchmark by latching on dataset-specific correlations, as opposed to learning more general skills required in VQA.
We fine-tune a pretrained model on the train split of one benchmark (\eg, \gqa) and evaluate it on the validation split of a different one (\eg, \vg).
Overall, we evaluate models by fine-tuning them on each benchmark and testing them against all benchmarks.
If pretrained models are indeed learning the VQA skill, we expect to see a small drop in performance between the IID and OOD settings. 

The results are presented in \cref{fig:iid_ood}, with different evaluation benchmarks grouped on the $x$-axis. First, across all models and for each benchmark, we see a notable drop in the VQA accuracy from the IID to the OOD setting. While such a drop might be anticipated, we found the extent of the drop surprising given the impressive performance of current pretrained VL models.
For all models shown, the largest drops are observed when evaluating models on the \vizwiz benchmark.
Moreover, even the smallest performance drop, which happens when fine-tuning models on \vqavii and evaluating them on \vg, remains relatively large (\ie, $5.3$ points for \galbef).
These results show that \emph{pretrained models are largely learning the fine-tuning benchmark without learning to solve the VQA task}.

Second, we observe that fine-tuning on \vqavii results in the lowest drop in IID to OOD performance across all conditions---the \vqavii bar (shown in blue in \cref{fig:iid_ood}) is the closet to the IID one for \gqa, \vg, and \vizwiz.
We conclude that fine-tuning on \vqavii yields a model that best generalizes to OOD settings in our benchmarks. This result is not simply due to the size of the fine-tuning benchmark as \vg is larger than \vqavii. 
Similarly, all the models achieve highest OOD performance on \vqavii.
We conjecture that \vqavii is the most diverse benchmark of our selection.

\vspace{-0.2mm}
\subsection{Evaluating on Shared Answer Sets} \label{subsec:shared_ans_set_eval}

\begin{figure*}[t]
\centering
    \includegraphics[width=\textwidth]{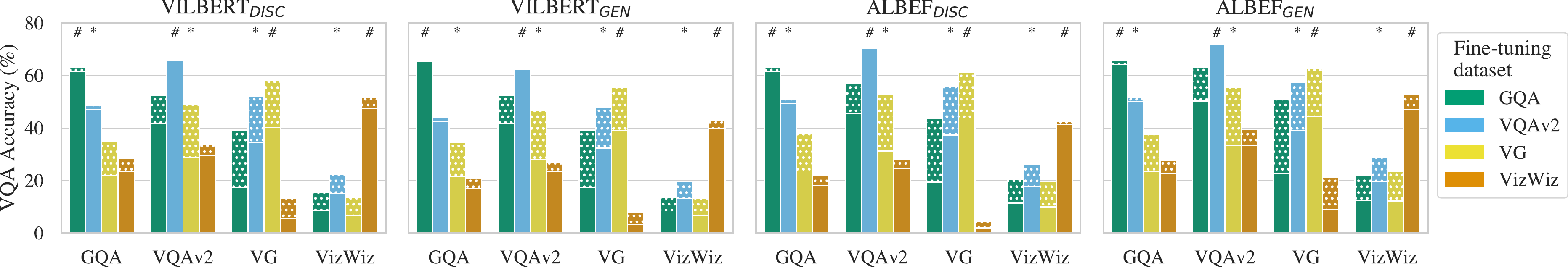}
\caption{
IID (\#) vs OOD performance when controlling for the shared shared answer set. Solid bars are as in \cref{fig:iid_ood}; stacked dotted bars are improvements when evaluating on questions with shared answer sets between IID and OOD settings. For IID, the shared answer set is computed with respect to a dataset denoted with *.}
\vspace{-.3cm}
\label{fig:in_vocab}
\end{figure*}

Discriminative models treat VQA as a multi-answer classification task over the set of \topk{} most frequent answers in the fine-tuning data.
This limits their performance: if a certain answer is not frequent in the fine-tuning data, a discriminative model will perform poorly for such an answer during test time.
While this limitation also affects IID evaluation, we expect it to have a stronger effect in OOD generalization (due to potentially different answer distributions between the fine-tuning and test sets).
We next examine to what extent this limitation affects OOD performance by controlling for the mismatch in answer sets between the fine-tuning and test sets. 
We do so by considering only the test questions whose answers are included in the \topk{} answers of a given fine-tuning dataset (for more details, see \cref{app:shared_answer_sets}).

\cref{fig:in_vocab} shows the improvement in the VQA accuracy over the IID and OOD evaluation accuracy (in \cref{fig:iid_ood}) when controlling for the shared answer set.
For IID evaluation, only one intersection of answer sets is reported, corresponding to the smallest gap between IID and OOD evaluation, with remaining numbers reported in \cref{tab:in_vocab_all} (\cref{app:more_results}).
Thus, the difference between the height of the IID bar (\#) and the OOD bar (*) with respect to which answer intersection between IID and OOD is computed, represents the best case scenario for OOD generalization, \ie, the least drop from IID to OOD.

We observe a similar pattern across the models: in most cases, using a shared answer set improves the performance.
\emph{Overall, we still observe a notable gap between the OOD and IID settings for the best case OOD generalization scenario, showing that a shared answer set does not circumvent the difficulty of OOD generalization for these models.}
A few cases where IID evaluations with a shared answer set hurt performance are discussed in \cref{app:more_results}.
When evaluating on the shared answer set, we further examine if the drop in accuracy from IID to OOD is due to the low frequency of the test answer classes in the OOD fine-tuning set. The details of the correlation computation and the results are explained in \cref{app:ans_freq_corr} and \cref{tab:acc_freq_corr_albef_gen}, respectively.
This result indicates that frequency of the answer class is a contributing factor to the weak OOD generalization, but we also explore other causes in \cref{sec:qual_study}.

\begin{table}[t]
     \setlength{\tabcolsep}{3.5pt}
     \centering
     \small
     \begin{tabular}{@{}l@{\hspace{4\tabcolsep}}cccc}
         \toprule
                 & \vqavii & \gqa & \vg & \vizwiz \\\midrule
         \vqavii     & 92.9	& 96.7       	& 65.1  & 43.6 \\
         \gqa     & 73.5	& 99.9 & 44.8        & 36.6 \\
         \vg      & 52.7	& 62.4 & 74.2  & 32.3 \\
         \vizwiz	& 79.4	& 82.5 & 40.9 & 86.2 \\
         \bottomrule
     \end{tabular}
     \caption{Maximum achievable accuracy for all test answers based on the \topk{} answers present in the respective fine-tuning sets. Rows correspond to fine-tuning datasets, columns correspond to the test benchmarks}
     \label{tab:max_score_overall_vilbert}
     \vspace{-0.4cm}
\end{table}

\subsection{The Case for the Generative Evaluation} \label{subsec:gen_eval}

A discriminative model cannot correctly answer questions for which the answers lie outside the predefined \topk{} classes; therefore, by treating VQA as a classification task, we can define the upper-bound performance of discriminative models on VQA by computing the accuracy given all answers in the test set being answered correctly.
The upper-bound VQA accuracy is shown in \cref{tab:max_score_overall_vilbert}; we observe a large drop from IID to OOD evaluations for most conditions. 
\vizwiz has the lowest achievable accuracies in OOD evaluation.

\begin{figure*}[t]
\centering
    \includegraphics[width=\textwidth]{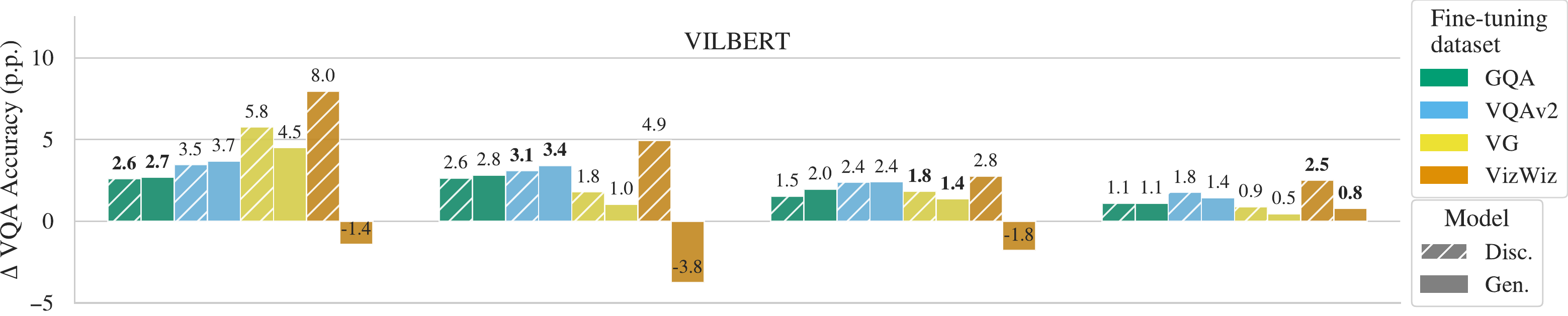}
    \includegraphics[width=0.9\textwidth, left]{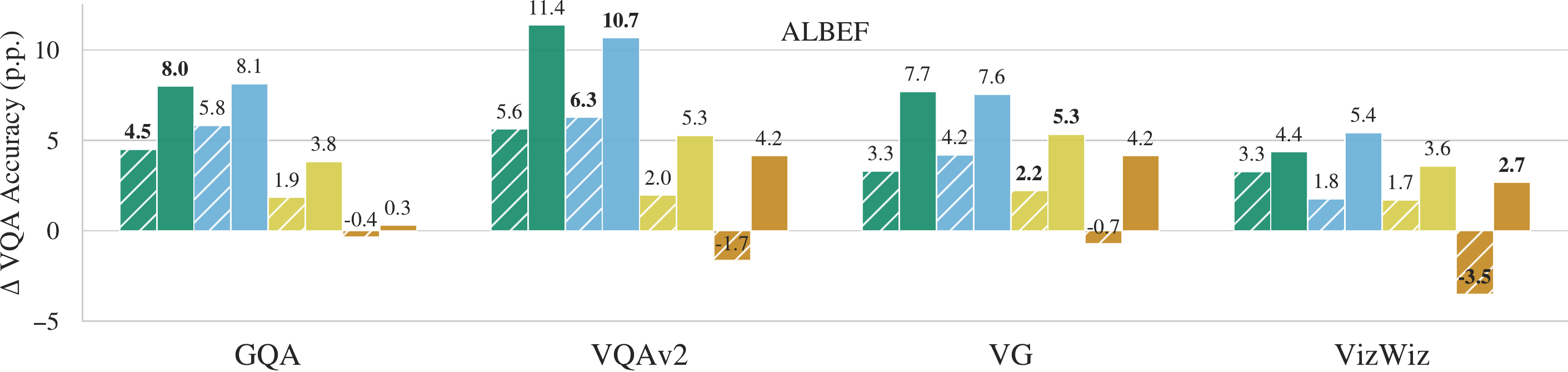}    
\caption{
Percentage point difference in VQA accuracy between models with and without multimodal pretraining, for OOD and IID (highlighted in bold) evaluations.  All models are initialized with BERT weights.}
\vspace{-0.2cm}
\label{fig:pretraining_effect}
\end{figure*}

\begin{figure}[t!]
\centering
    \includegraphics[scale=.4]{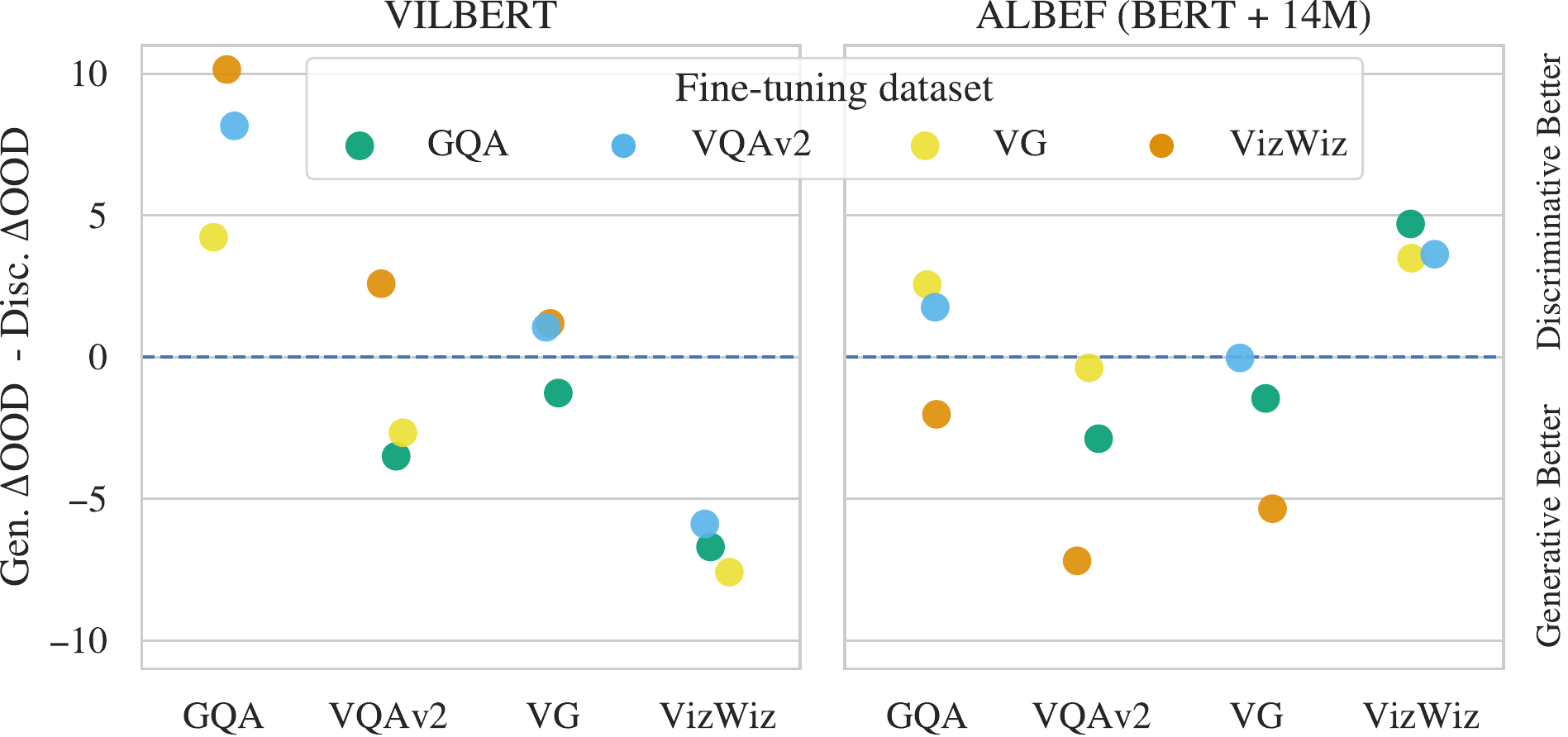}%
\vspace{-.15cm}
\caption{Difference in $\Delta\,\text{OOD}$ values between discriminative and generative models.}
\label{fig:iid_ood_gen_dis_gap}
\vspace{-.7cm}
\end{figure}

However, our \dalbef and \dvilbert models still perform notably worse than maximum achievable accuracy in all settings (smallest gap of 21.5\% across all conditions, see \cref{fig:diff_iid_ood} in \cref{app:more_results}); \emph{as a result, the poor OOD performance in the discriminative setting is not simply due to the low maximum achievable accuracy}.
We conclude that the common practice of modeling VQA as a classification task severely limits the generalization capability of models to new datasets. On the other hand, generative models do not suffer from a fixed class set. They can generate a larger set of answers---all words for which the tokens occur in the pretraining data, including those that are out-of-vocabulary for the given VQA fine-tune datasets.
We argue that generative modeling is a more promising solution for real-world application of VQA; similarly, recent work has identified text generation as a way to unify various \vl{} tasks \citep[\eg,][]{VL-T5, wang2022simvlm, alayrac2022flamingo}.

We next ask \emph{whether our \gvilbert and \galbef models are more successful in OOD generalization compared to their discriminative counterparts}. For each model (\ie, generative/discriminative \albef/\vilbert), we first calculate the gap between the IID setting and each OOD setting (\ie, $\Delta\,\text{OOD}$), resulting in three values per benchmark. For instance, for the \vqavii benchmark, $\Delta\,\text{OOD}$ numbers are calculated between the model fine-tuned on \vqavii and those fined-tuned on \vg, \gqa, and \vizwiz. Note that the higher the $\Delta\,\text{OOD}$ value, the poorer a model is in OOD generalization. We then compute the difference between the $\Delta\,\text{OOD}$ values of the generative and discriminate models.  \cref{fig:iid_ood_gen_dis_gap} visualizes this result; the benchmarks are shown on the $x$-axis and each circle represents the difference in $\Delta\,\text{OOD}$ values between the generative and discriminative model for a given fine-tuning dataset. If a generative model is more robust to OOD evaluation, we expect to see smaller $\Delta\,\text{OOD}$ value for that model compared to its discriminative counter part: when the circles are below the $x$-axis (depicting negative values), the generative model is more robust than the discriminative one. We observe \galbef models often outperform their discriminative counterparts with respect to OOD generalization.

\section{The Effect of Multimodal Pretraining }
\label{sec:pretrain}

Previous work has shown that pretraining on multimodal (\ie, image--text)  data improves IID performance \citep[\eg,][]{lu2019vilbert, li2021align}; here, we ask if multimodal pretraining can help in OOD settings as well.
We repeat the experiments in \cref{sec:ood_gen} without pretraining our models on multimodal data; instead we train the models on the train split of one benchmark and test it on the validation split of another.
\cref{fig:pretraining_effect} shows the difference between the VQA accuracy of models with and without multimodal pretraining: each bar shows the gap between a bar in \cref{fig:iid_ood} and the equivalent experiment without multimodal pretraining.

We observe that multimodal pretraining is helpful in almost all conditions, since the majority of values displayed in  \cref{fig:pretraining_effect} are positive.
Pretraining is improving OOD performance likely because it can reduce the gap between the train and OOD test data by potentially exposing the model to a more diverse set of data points during pretraining. In our experiments, the maximum gain from multimodal pretraining is indeed observed in OOD settings for both \vilbert  (fine-tune on \vizwiz; test on \gqa) and \albef (fine-tune on \gqa ; test on \vqavii); however,  \emph{multimodal pretraining is \textbf{not} always more useful in OOD settings compared to IID ones}. For example, when evaluating \vilbert on \vqavii, pretraining helps the IID setting more than some of the OOD ones. 
Lastly, multimodal pretraining is detrimental for some cases where models are fine-tuned on \vizwiz.

We observe that multimodal pretraining is more effective for the generative \albef compared to the discriminative \albef (cf. the shaded and solid bar with the same color in \cref{fig:pretraining_effect} bottom).
For the \vilbert model, we generally do not observe such a pattern---discriminative and generative models mostly show comparable improvements due to multimodal pretraining.
We observe only small improvements when increasing the size of the multimodal pretraining dataset for the \albef model (see \cref{fig:pretraining_effect_albef_small} in \cref{app:more_results} for more details).

\begin{table}[t!]
    \vspace{-2pt}
    \setlength{\tabcolsep}{3.5pt}
    {\small
    \begin{center}
        \begin{tabular}{@{}l@{}c@{\hspace{\tabcolsep}}ccc}
            \toprule
            \textbf{Model} & \textbf{MM PT} & \textbf{VQAv2} & \textbf{VQA-CP} & \textbf{Drop}\\
            \midrule
            \textcolor{gray}{CF-VQA} & \textcolor{gray}{--} & \textcolor{gray}{53.6} & \textcolor{gray}{63.5} & \textcolor{gray}{9.9} \\
            \midrule
            \dvilbert & no & 66.7 & 42.5 & 24.2 \\
            \dvilbert & yes & 67.0 & 42.9 & 24.1 \\
            \midrule
            \dalbef & no & 64.0 & 40.1 & 23.9 \\
            \dalbef & yes (4M) & 70.0 & 44.4 & 25.6 \\
            \dalbef & yes (14M) & 70.3 & 45.2 & 25.1 \\
            \midrule
            \galbef & no & 61.4 & 36.6 & 24.8 \\
            \galbef & yes (4M) & 71.0 & 49.2 & 21.8 \\
            \galbef & yes (14M) & 72.1 & 49.6 & 22.5\\
            \bottomrule
        \end{tabular}
    \end{center}
    }
\vspace{-10pt}
\caption{Performance of models on \vqavii (IID) and \vqacp (OOD). The last column shows drop in performance from \vqavii to \vqacp. MM PT: Multimodal Pretraining.}
\vspace{-8pt}
\label{tab:vqa-cp}
\end{table}

\section{Evaluation on VQA-CP}
\label{sec:vqacp}

In this section, we evaluate the models\footnote{\albef and \dvilbert (using the official codebase).}
on the VQA under Changing Priors dataset (\vqacp;~\citealt{Agrawal_2018_CVPR}). This dataset is designed such that, for every question type, train and test splits have different prior distributions of answers. Thus, models that overfit to answer priors in training data and lack sufficient visual grounding show poor generalization on the \vqacp test set.
For comparison, we also evaluate the performance of Counterfactual VQA (CF-VQA; \citealt{niu2021counterfactual}), a state-of-art method on \vqacp, which does not use either the Transformer architecture nor multimodal pretraining.
However, it explicitly tackles the language (\ie, question and answer) biases in VQA-CP.

\cref{tab:vqa-cp} shows that for all the Transformer-based models, there is a large drop in the performance (at least 22\%) from \vqavii to \vqacp. Thus, in spite of advances in the Transformer architecture and pretraining on diverse datasets, models are still overfitting to answer priors in the training data and lack sufficient visual grounding~\cite{Agrawal_2018_CVPR}. However, the drop is much less for CF-VQA (10\%), suggesting that \emph{incorporating inductive biases specific to the generalization problem} (\ie, modeling language bias) \emph{helps more than the Transformer architecture or scaling up the amount of pretraining data}.
We also observe that the drop from \vqavii to \vqacp is often lower for the generative \albef than the discriminative \albef (except for \albef without any multimodal pretraining). Thus, \emph{generative models are more robust than the discriminative ones}, especially when they are pretrained (similarly to the observations made in \cref{subsec:gen_eval}). 
As for the effect of pretraining, for generative \albef, pretraining helps reduce the drop from \vqavii to \vqacp. However, for discriminative models, pretraining does not seem to help generalization (in fact, it worsen \albef).

\section{Qualitative Analysis}
\label{sec:qual_study}
To dig deeper into the potential causes of the poor OOD generalization of our pretrained models, we perform a qualitative study.
To this end, we randomly sample and manually examine failure cases in top-30 answer classes with the highest performance drop when moving from IID to OOD evaluation. We only focus on answer classes that are present in both the train and test splits, ensuring that performance drop is not due to the absence of answer classes in the training set.
We report the top-5 classes that contribute the most to the drop in performance for each OOD setting in \cref{tab:top_ans_class_all} (\cref{app:qual_study}). We notice that the following answer classes appear frequently across different OOD settings: yes/no answers, directions (left/right), colors, and numbers. In the following, we discuss a few major potential causes for the poor OOD generalization, and mention \dvilbert responses as examples in the discussion, although similar observations hold for other models.

\textbf{Poor reasoning skills.} 
Models evaluated on \gqa, but fine-tuned on another dataset, show the highest performance drop on classes such as ``yes'', ``no'', ``right'', ``left'', ``top'', and ``bottom''. For instance, \dvilbert fine-tuned on \vqavii, and evaluated on \gqa underperforms \dvilbert that has been both fine-tuned and evaluated on \gqa by 24\% for the answer class ``no.'' 
Upon qualitative examination, we find that for many of such failure cases, the \gqa questions are more compositional and hence require more complex reasoning (\eg, ``Are there both bison and zebras in the image?'', ``Is the cheese to the right or to the left of the empty plate?'') than the questions for the same answer classes in other datasets (\eg, from \vqavii train set: ``Is the TV turned on?'', ``Which hand is the man holding up?''). 
This study re-affirms previous findings that VQA models lack sufficient logical, spatial, and compositional reasoning skills \cite{Johnson_2017_CVPR,Hudson_2019_CVPR} but for the more recent, pretrained Transformer models.

\begin{figure}[t]
    \centering
    \includegraphics[width=0.48\textwidth, trim={0cm 5.2cm 5.5cm 0cm}, clip]{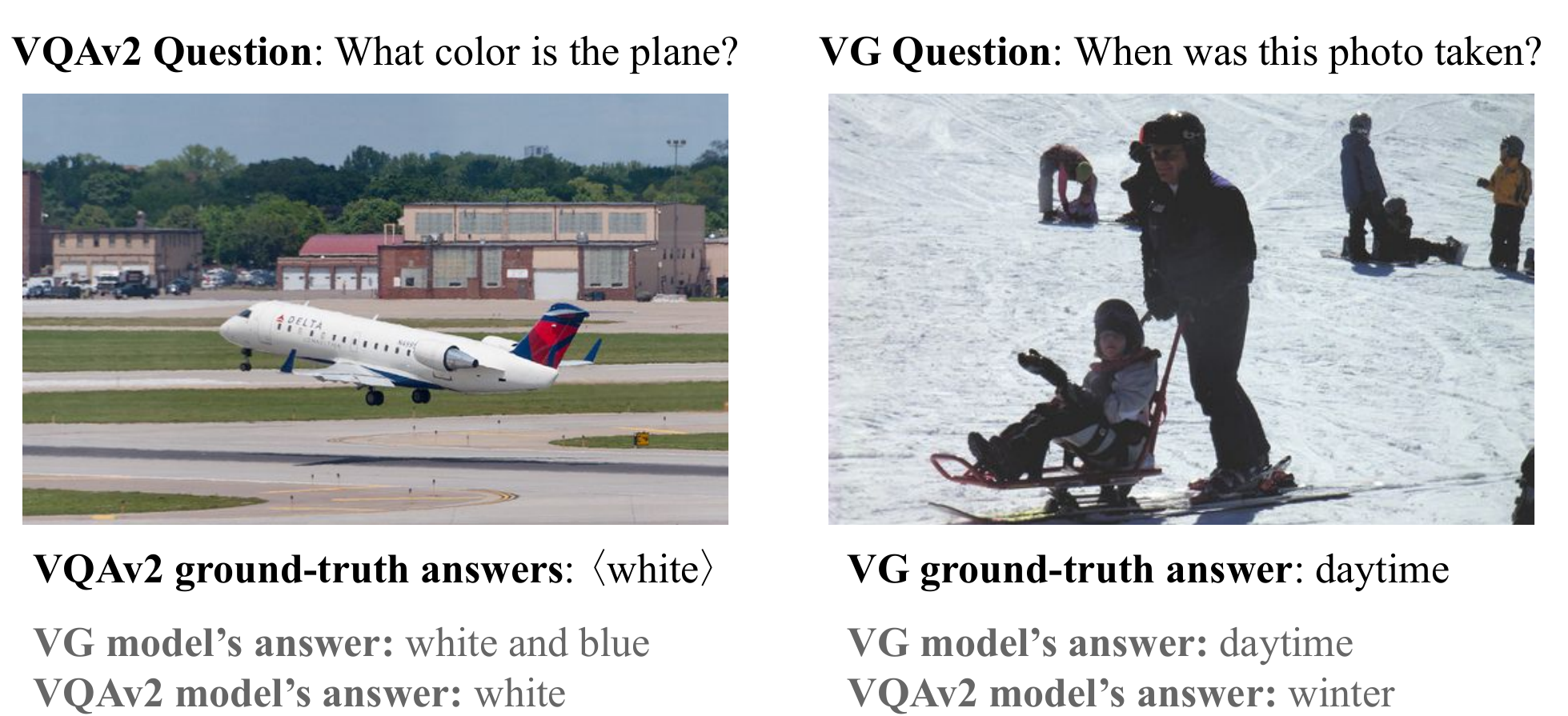}
    \caption{Examples where models' prediction are correct but not accounted for in the ground-truth set. $\langle~\rangle$ denotes a list of unique (out of 10) ground-truth answers. \vg (\vqavii) model refers to a \dvilbert fine-tuned on \vg (\vqavii).}
    \label{fig:strict_eval_metric}
    \vspace{-15pt}
\end{figure}

\textbf{Overfitting to the answer priors.} Previous studies have shown that VQA models tend to be biased towards the prior distribution of answers in the training set (per question type) \cite{Agrawal_2018_CVPR}. We find that this limitation exists in the more recent pretrained models as well, and it is especially hurtful in the OOD settings because the priors need not be the same across train and test sets, unlike in the IID settings. For instance, \dvilbert fine-tuned on \vqavii predicts ``2'' for a lot questions with target answer ``1'' in the \vg test set. Similarly, sometimes \dvilbert fine-tuned on \vg incorrectly predicts ``helmet'' for \vqavii test questions such as ``What is the skateboarder wearing to protect his head?'', ``What protective gear is he wearing?'' when the skateboarder is not wearing anything. This indicates that the model is relying on answer priors rather than visual grounding. Our experimental results on VQA-CP (\cref{sec:vqacp}) directly quantify the extent of such limitations in current models.

\textbf{Overfitting to the question format.} We observe instances of models failing to correctly answer questions  when the format of the questions changes between the fine-tuning and test sets.
For instance, questions about ``chair'' in the \vqavii fine-tuning set are mostly of the form ``What is \dots{} sitting on?'' whereas in the \gqa test set, they are mostly of the form ``What kind of furniture is \dots?''. Thus, the ``chair'' class accuracy of \dvilbert fine-tuned on \vqavii drops from 48\% when tested on \vqavii to 38\% on the \gqa test set. Similarly, \dvilbert fine-tuned on \gqa fails terribly for ``dog'' and ``cat'' classes on the \vg test set (accuracy drops of 47\% and 43\% respectively between \gqa--\gqa (fine-tuned on \gqa, tested on \gqa) and \gqa--\vg). \gqa questions are mostly of the form ``What animal \dots?'' or ``What kind of animal \dots?'' whereas \vg questions often do not mention the word ``animal'' and are of the form ``Who is \dots?'' or ``What is \dots?'' (\eg, ``Who is holding the Frisbee?'', ``What is on the leash?''). To the best of our knowledge, no previous work has reported such behavior of VQA models (i.e., they tend to overfit to the question format). 

Finally, we observe cases where correct model responses are evaluated as incorrect by the VQA evaluation metric, as such responses differ from the ground-truth answers. In the next section, we provide examples of such cases and examine the impact of \textbf{stringent evaluation metric} on poor OOD generalization by engaging human raters to evaluate responses.

\section{Human Evaluation}
\label{sec:human_eval}

In our qualitative study, we observed that the stringent nature of the standard VQA evaluation metrics (\ie, performing string matching of model responses with a small set of ground-truth answers) repeatedly penalizes models for correct responses because those responses do not exist in the set of ground-truth answers (\cref{fig:strict_eval_metric}).
For example, the evaluation metric fails to take into account differences (between model response and ground-truth) due to specificity of the answers (\eg, ``on table'' vs. ``table'', ``pizza slices'' vs. ``pizza''), synonyms, and different interpretations of the question (\eg, \cref{fig:strict_eval_metric} right).

In this section, we aim to quantify how robust standard VQA metrics are by performing human evaluation of our models for both IID and OOD settings. The details of the setup and in-depth results are provided in \cref{app:human_eval}. Below we present our main findings.

\begin{figure}[t!]
\centering
    \includegraphics[width=0.47\textwidth]{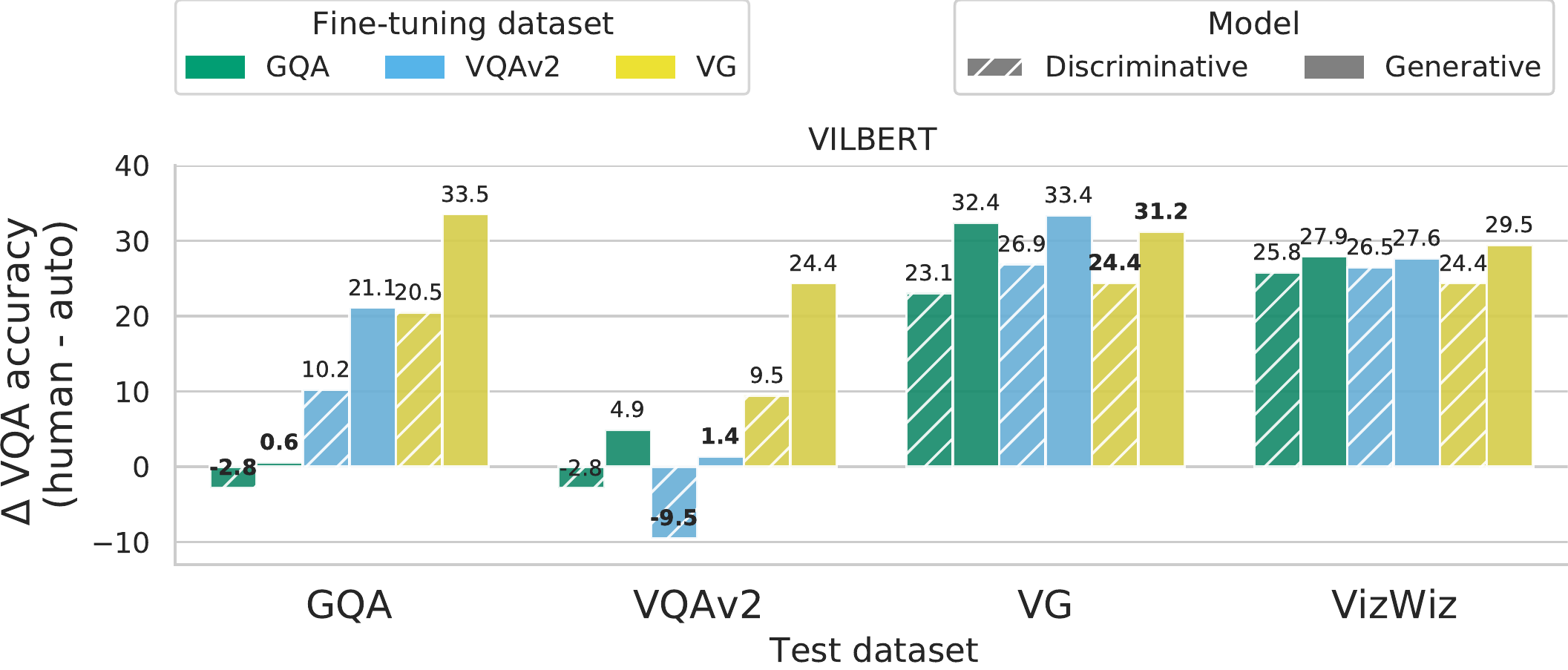}
\caption{Difference in human and automatic accuracy of \dvilbert (shaded bars) and \gvilbert (plain bars) for \gqa, \vqavii, \vg and \vizwiz. Accuracies in bold denote the IID settings.}
\label{fig:human_results_subset}
\vspace{-10pt}
\end{figure}

\paragraph{Human evaluation yields notably higher accuracies than the automatic evaluation.} 
This is shown in \cref{fig:human_results_subset}, where the increase can be up to 33.5\% when moving from automatic to human evaluation.\footnote{In some cases, human evaluation yields lower accuracy than the automatic evaluation. We discuss this in \cref{app:human_eval}.}
This implies the current automatic metrics miss out on a lot of correct responses due to their stringent nature.
Interestingly, this increase in model accuracy from automatic to human evaluation is higher for \gvilbert than \dvilbert for all the benchmarks. This is expected because the generative model is more likely to produce longer, more varied answers, which might not be awarded using automatic metric but are still correct responses. Moreover, human evaluation helps OOD settings more than the IID settings for most of the benchmarks (\eg, \gqa, \vqavii). This is also expected, because in the OOD settings, a model might not learn the format of the test answer (``on table'' vs. ``table'', ``clear vs. sunny'') from the train set (unlike in the IID settings) and hence it is more likely to be penalized by the automatic accuracy metric.
Thus, we conclude that the currently used accuracy metrics for VQA are not robust, especially for generative models and OOD evaluation settings. Hence, to more accurately evaluate the goodness of our models, \emph{we need to develop better evaluation metrics for VQA}.
     
\paragraph{Even after human evaluation, models still exhibit poor OOD generalization.}
Although human evaluation improves the models' accuracies and more so for the OOD than the IID settings, we observe that the models' performance in OOD settings is still worse compared to that of IID settings, albeit with reduced margin (see \cref{app:human_eval} for quantitative results).
We also note that while \dvilbert usually outperformed \gvilbert with the automatic evaluation, \gvilbert outperforms \dvilbert for all the test sets under human evaluation. This reinforces the observations in \cref{subsec:gen_eval} regarding stronger OOD generalization of generative models over discriminative ones.

\section{Conclusion}
\label{sec:conclusion}

In this study, we show that, despite their impressive performance when evaluated on test data drawn from the same distribution as the training data, recent \vl{} models perform poorly in out-of-distribution (OOD) settings.
We conclude that these models learn to solve specific benchmarks as opposed to the skill of visual question answering (VQA).
Interestingly, in most cases, we observe that the generative models are more robust to OOD generalization compared to the discriminative ones. Moreover, pretraining the models on large image--text data often helps in OOD generalization.
Our results also highlight the importance of human evaluation for a more accurate assessment of model performance: we find that the current VQA automatic metrics miss out on a notable number of correct model responses.
Human evaluation is especially important as the community is shifting towards generative VQA models which, unlike discriminative ones, can produce answers that go beyond those seen in a training/fine-tuning dataset.
Finally, to make progress towards more capable models, we need more rigorous evaluation protocols that shed light on models' strengths and short-comings. We believe testing models in OOD settings is a step towards this direction as it helps evaluate models for general skills required to solve the task as opposed to benchmark-specific correlations.

\section*{Limitations}
We list some limitations of our work which could benefit from future investigations.

First, when exploring potential factors for poor OOD generalization, our quantitative analysis focused only on differences in answer distributions between fine-tuning and test datasets. However, future work should investigate differences in question distribution, image distribution and combinations of these three variables.

Second, it would be interesting to conduct further investigation to understand why multimodal pretraining does not help in certain cases. A correlation analysis between improvement in accuracy (due to multimodal pretraining), and between pretraining and fine-tuning/test data could be useful.

Third, the models investigated in our study (\vilbert and \albef) are pretrained on a relatively small number (millions) of data points compared to language-only pretrained transformers, such as BERT, trained on billions of tokens. Such large-scale pretraining has been shown to improve OOD robustness for language-only models \cite{hendrycks-etal-2020-pretrained}. Hence, we leave for future work to investigate multimodal models trained on billions of image--text pairs (for instance, LAION-5B; \citealt{schuhmann2021laion}).

Lastly, in this study, we only focus on standard VQA evaluation metrics for each benchmark. However, it would be interesting to also evaluate the robustness of metrics such as WUPS \cite{NIPS2014_d516b136} that compute answer similarities based on the distance between them in the WordNet \cite{miller1995wordnet} tree and are expected to be more robust than the standard metrics.

\section*{Ethics Statement}
Below we present some considerations related to the ethical and broader impact of our work.

First, all datasets used in our study are from published work and are publicly available, including the \vizwiz{} data \cite{Gurari_2018_CVPR} which has been curated from visually impaired users and released publicly after proper filtering to preserve the privacy of the users.

Second, for human evaluation of our models, we collected human data via the Amazon Mechanical Turk platform. We detail the data collection process and measures taken to control the quality of collected data in \cref{app:human_eval}. As for the ethical considerations related to collecting data from human subjects, our data collection campaign was approved by an ethics review board in our institution.
Human subjects were paid at the rate of 0.15 USD per HIT (Human Intelligence Task) resulting in an hourly payment well above minimum wage.

Third, by testing models on a data distribution different from the training one, the OOD evaluation setting studied in our work has the following broader impacts: it highlights (1) the challenges of generalizing to real-world VQA datasets such as \vizwiz, and (2) the kind of biases learned (and also potentially amplified) by the models.

Lastly, we discuss both potentially beneficial and harmful applications of the task of Visual Question Answering studied in our work. VQA has many potential applications beneficial for society:
\begin{itemize}[topsep=2.5pt]
\setlength\itemsep{0.0001cm}
\item  Aiding  visually  impaired  users  in  understanding  their  surroundings  (Human: \texttt{What is on the shelf above the microwave?} AI: \texttt{Canned containers.})
\item  Teaching children through interactive demos (Kid: \texttt{What animal is that?} AI: \texttt{That is Dall Sheep. You can find those in Alaska.})
\item  Aiding analysts in processing large quantities of visual surveillance data (Analyst: \texttt{What kind of car did the man in red shirt leave in?} AI: \texttt{Blue Toyota Prius.})
\item  Interacting with in-home physical robots (Human: \texttt{Is  my laptop  in  my  bedroom  upstairs?}  AI: \texttt{Yes.}  Human: \texttt{Is  the  charger  plugged in?})
\item  Making visual social media content more accessible (AI: \texttt{Your friend Bob just uploaded  a  picture  from  his  Hawaii  trip.} Human: \texttt{Great,  is  he  at  the  beach?} AI: \texttt{No, on a mountain.})
\end{itemize}
But like most other technology, VQA could also be used for potentially harmful applications such as:
\begin{itemize}[topsep=2.5pt]
\setlength\itemsep{0.0001cm}
\item  Invasion of individual’s privacy by using VQA to query streams of video data being recorded by CCTV cameras at public places.
\item  Visually impaired users often need assistance with parsing data containing personal information~\cite{ahmed2015privacy}, such as credit cards, personal mails, etc. Such VQA systems could be configured to leak/retain personally identifiable information.
\end{itemize}

\section*{Contributions}
\emph{Aishwarya Agrawal} initiated and designed the project, ran experiments and analyses on the official codebase of ViLBERT, contributed significantly to paper writing, and provided project support and advice.
\emph{Ivana Kaji\'{c}} was responsible for the project's technical infrastructure for the re-implementation of ViLBERT, ran experiments and analyses, and contributed significantly to paper writing.
\emph{Emanuele Bugliarello} co-led the data preparation, ran experiments and analyses on ALBEF, and contributed significantly to paper writing.
\emph{Elnaz Davoodi} co-led the data preparation, and helped setting up and running re-implementation experiments.
\emph{Anita Gergely} led the human evaluation experiments, and contributed to paper writing.
\emph{Phil Blunsom} provided project advice.
\emph{Aida Nematzadeh} provided significant project support and advice, helped running experiments, and contributed significantly to paper writing.

\section*{Acknowledgements}
We are grateful to Antoine Miech, Lisa Anne Hendricks and Chris Dyer for their constructive feedback.  {\scriptsize\euflag}
During this project, \emph{Emanuele Bugliarello} was supported by the funding from the European Union's Horizon 2020 research and innovation programme under the Marie Sk\l{}odowska-Curie grant agreement No 801199.

\bibliographystyle{acl_natbib}
\bibliography{references}

\clearpage
\appendix
\section{Experimental Setup Details}
\label{app:exp_details}

In this section, we report additional details regarding our experimental setup.

\subsection{Models}
\label{app:models}

We evaluate the performance of two strong models, \vilbert~\cite{lu2019vilbert} and \albef~\cite{li2021align}.
These models belong to the family of pretrained Transformer that has recently achieved state-of-the-art performance on several \vl{} tasks, and are specifically instances of dual-stream architectures~\cite{bugliarello-etal-2021-multimodal}.
In this paradigm, models are first pretrained on a large collection of image--caption pairs, and then fine-tuned to solve specific downstream tasks.
\vilbert is pretrained using three objectives, masked language modeling (MLM;~\citealt{BERT}), masked region modeling and image--text matching (ITM;~\citealt{chen2019uniter}).
\albef is pretrained using MLM, ITM and an image--text contrastive loss~\cite{li2021align}.
We refer the reader to \cref{sec:exp_setup} for an overall description of these models.
\cref{tab:models_specs} lists pretraining and architecture details for both models.
All the models were fine-tuned using the AdamW~\cite{loshchilov2018decoupled} optimizer, with model-specific hyperparameters in \cref{tab:hyper_all}.

\paragraph{\textsc{ViLBERT}.}
In this model, the textual inputs are first processed through 6 Transformer layers, before being combined with visual inputs through inter- and intra-modal attention layers.
We re-implement this architecture, and confirm comparable performance by reproducing its results with the ones obtained through the official codebase\footnote{\url{https://github.com/facebookresearch/vilbert-multi-task/}.} (see \cref{tab:vilbert_vs} for IID performance of both implementations).
A key difference between the two implementations is in the image features: while the official model uses 10--100 regions of interest (RoIs) from a ResNeXt-152~\cite{Xie_2017_CVPR}, our re-implementation relies on 100 RoIs extracted by Faster R-CNN~\cite{ren2015faster} trained on \vg.

We then extend our codebase to implement a generative version of \vilbert by replacing the discriminative decoder with an autoregressive decoding head.\footnote{Our implementation of the generative decoder follows that of \texttt{TransformerDecoder} available at \url{https://github.com/pytorch/pytorch/blob/master/torch/nn/modules/transformer.py}.}
The decoder is trained with teacher-forcing, and in datasets with several responses, the most frequent response is used as the ground truth response. Pretraining on CC, when used, is done by generating text. We also examine the effect of pretraining on both the encoder and decoder, and find the learning to be more stable when using only pretrained encoder, although further hyperparameter exploration could mitigate this difference. We study the effects of multimodal pretraining on the Conceptual Captions dataset~\cite{sharma-etal-2018-conceptual} with 3M images.

\paragraph{\textsc{ALBEF}.}
Like \vilbert, \albef is a dual-stream encoder but with two main differences: first, the visual inputs are image patches that are processed through a vision Transformer~\cite{dosovitskiy2021an}; and second, the cross-modal interactions happen through standard Transformer cross-attention at each layer (whereas \vilbert uses co-attention layers specifically designed for intra- and inter-modal interactions).
In addition, the model is trained with pseudo-targets that are generated from a moving-average version of its weights.
We run experiments using the official codebase.\footnote{\url{https://github.com/salesforce/ALBEF}.}
The visual backbone is a DeIT-B/16~\cite{touvron2021training} pretrained on ImageNet-1k~\cite{5206848} at resolution 224$\times$224, and further trained during the multimodal pretraining phase.
For the downstream VQA benchmarks, we follow the authors and resize input images to 384$\times$384 and apply random augmentation during fine-tuning. 
\citet{li2021align} formulated the VQA task as generative by adding a 6-layer Transformer decoder initialized from the pretrained encoder.
We follow this approach and also evaluate a discriminative version by learning a two-layer MLP with ReLU~\cite{agarap2018deep} non-linearity in between, following the authors' setup for the Visual Entailment benchmark~\cite{xie2019visual}.
We found the hyperparameters proposed by~\citet{bugliarello-etal-2021-multimodal} to work better.
During inference, we evaluate \albef in two ways: first, following the authors, we rank the in-domain candidate answers based on their likelihood; second, we let the model generate any possible answer in an open-ended fashion through greedy decoding.
We found these two approaches to minimally affect final performance (see~\cref{tab:rank_vs_gen_albef}). 
Unless otherwise specified, we report results given by generation as it reflects open-ended question answering.

\begin{table}[t]
\vspace{-2pt}
\setlength{\tabcolsep}{3.5pt}
{\small
\begin{center}
    \resizebox{0.48\textwidth}{!}{
        \begin{tabular}{@{}l@{\hspace{4\tabcolsep}}r|lrr}
        \toprule
        \textbf{Model} & \textbf{\# Params} & \textbf{Pretrain data} & \textbf{\# Images} & \textbf{\# Captions} \\
        \midrule
        \vilbert & 240M & CC & 3.3M & 3.3M \\
        \albef (4M) & 450M & ~~+COCO+SBU+VG & 4M & 5M \\
        \albef (14M) & 450M & ~~~~+C12M & 14M & 15M \\
        \bottomrule
        \end{tabular}
    }
\end{center}
}
\vspace{-10pt}
\caption{Pretrained models details. ALBEF size includes both the main model and its moving average. Pretraining data: CC~\cite{sharma-etal-2018-conceptual}, COCO~\cite{lin2014microsoft}, SBU~\cite{ordonez2011im2text}, VG~\cite{krishna2017visual}, C12M~\cite{changpinyo2021conceptual}.}
\label{tab:models_specs}
\end{table}

\begin{table}[t]
\vspace{-2pt}
\setlength{\tabcolsep}{3.5pt}
{\small
\begin{center}
    \resizebox{0.48\textwidth}{!}{
        \begin{tabular}{@{}l@{\hspace{4\tabcolsep}}rrrr}
        \toprule
        \textbf{Model} & \textbf{\vqavii} & \textbf{\gqa} & \textbf{\vg} & \textbf{\vizwiz} \\
        \midrule
        Official codebase & 67.04 & 66.78 & 40.69 & 44.46 \\
        Re-implementation & 65.75 & 61.51 & 40.31 & 47.46 \\
        \bottomrule
        \end{tabular}
    }
\end{center}
}
\vspace{-10pt}
\caption{Comparison between the official and our codebases for \dvilbert in the IID setting.} \label{tab:vilbert_vs}
\vspace{-0.4cm}
\end{table}

\begin{table}[t]
\vspace{-2pt}
\setlength{\tabcolsep}{3.5pt}
\begin{subtable}[t]{0.48\textwidth}
    {\small
    \begin{center}
        \resizebox{\textwidth}{!}{
            \begin{tabular}{@{}l@{\hspace{4\tabcolsep}}rrrrr}
            \toprule
            \textbf{Benchmark} & \textbf{Qn len} & \textbf{\# Classes} & \textbf{BS} & \textbf{LR} & \textbf{\# Epochs} \\
            \midrule
            \vqavii & 16 & 3{,}129 & 256 & 4e-5 & 20 \\
            \gqa & 26 & 1{,}533 & 256 & 4e-5 & 20 \\
            \vg & 16 & 3{,}449 & 256 & 4e-5 & 20 \\
            \vizwiz & 40 & 3{,}112 & 256 & 4e-5 & 20 \\
            \bottomrule
            \end{tabular}
        }
    \end{center}
    }
    \vspace{-6pt}
    \caption{Parameters used for \vilbert models. The internal codebase uses LAMB optimizer with the initial LR of 1e-3, with the best checkpoint selected on eval dataset.}
    \label{tab:hyper_off_dvilbert}
    \vspace{0.3cm}
\end{subtable}

\begin{subtable}[t]{0.48\textwidth}
    {\small
    \begin{center}
        \resizebox{\textwidth}{!}{
            \begin{tabular}{@{}l@{\hspace{4\tabcolsep}}rrrrr}
            \toprule
            \textbf{Benchmark} & \textbf{Qn len} & \textbf{\# Classes} & \textbf{BS} & \textbf{LR} & \textbf{\# Epochs} \\
            \midrule
            \vqavii & 16 & 3{,}129 & 256 & 1e-4 & 20 \\
            \gqa & 26 & 1{,}533 & 256 & 1e-4 & 20 \\
            \vg & 16 & 3{,}449 & 256 & 1e-4 & 20 \\
            \vizwiz & 40 & 3{,}129 & 256 & 1e-4 & 20 \\
            \bottomrule
            \end{tabular}
        }
    \end{center}
    }
    \vspace{-6pt}
    \caption{Discriminative \albef.}
    \label{tab:hyper_ours_dvilbert}
    \vspace{0.3cm}
\end{subtable}
\begin{subtable}[t]{0.48\textwidth}
    {\small
    \begin{center}
        \resizebox{\textwidth}{!}{
            \begin{tabular}{@{}l@{\hspace{4\tabcolsep}}rrrrr}
            \toprule
            \textbf{Benchmark} & \textbf{Qn len} & \textbf{Answer len} & \textbf{BS} & \textbf{LR} & \textbf{\# Epochs} \\
            \midrule
            \vqavii & 16 & 6 & 256 & 2e-5 & 20 \\
            \gqa & 26 & 5 & 256 & 2e-5 & 20 \\
            \vg & 16 & 8 & 256 & 2e-5 & 20 \\
            \vizwiz & 40 & 11 & 256 & 2e-5 & 20 \\
            \bottomrule
            \end{tabular}
        }
    \end{center}
    }
    \vspace{-6pt}
    \caption{Generative \albef.}
    \label{tab:hyper_ours_dvilbert}
    \vspace{0.0cm}
\end{subtable}
\caption{Hyperparameters used in our experiments. Question and answer lengths are in tokens, BS is the batch size, LR is the learning rate.}
\label{tab:hyper_all}
\vspace{-0.4cm}
\end{table}

\begin{table}[t]
\vspace{-2pt}
\setlength{\tabcolsep}{3.5pt}
\begin{subtable}[t]{0.48\textwidth}
    {\small
    \begin{center}
            \begin{tabular}{@{}l@{\hspace{4\tabcolsep}}rrrr}
            \toprule
            & \textbf{\vqavii} & \textbf{\gqa} & \textbf{\vg} & \textbf{\vizwiz} \\
            \midrule
            \textbf{\vqavii} & 72.37 & 50.56 & 38.94 & 19.81 \\
            \textbf{\gqa} & 50.32 & 64.26 & 22.80 & 12.51 \\
            \textbf{\vg} & 33.40 & 24.99 & 43.35 & 12.60 \\
            \textbf{\vizwiz} & 34.17 & 22.73 & 8.89 & 48.44 \\
            \bottomrule
            \end{tabular}
            
    \end{center}
    }
    \vspace{-6pt}
    \caption{Ranking-based evaluation of \galbef.}
    \label{tab:rank_galbef}
    \vspace{0.3cm}
\end{subtable}
\begin{subtable}[t]{0.48\textwidth}
    {\small
    \begin{center}
            \begin{tabular}{@{}l@{\hspace{4\tabcolsep}}rrrr}
            \toprule
            & \textbf{\vqavii} & \textbf{\gqa} & \textbf{\vg} & \textbf{\vizwiz} \\
            \midrule
            \textbf{\vqavii} & 72.09 & 50.10 & 39.20 & 19.81 \\
            \textbf{\gqa} & 50.33 & 64.24 & 22.79 & 12.50 \\
            \textbf{\vg} & 33.39 & 23.64 & 44.55 & 12.25 \\
            \textbf{\vizwiz} & 34.44 & 22.80 & 9.13 & 47.14 \\
            \bottomrule
            \end{tabular}
    \end{center}
    }
    \vspace{-6pt}
    \caption{Generation-based evaluation of \galbef.}
    \label{tab:gen_galbef.}
    \vspace{0.0cm}
\end{subtable}
\caption{Performance of \galbef (14M) when tested via ranking (top) and generation (bottom). Rows correspond to the fine-tuning datasets, columns correspond to the test benchmarks. The model performs similarly in both setups. We found similar results for \galbef (4M).} \label{tab:rank_vs_gen_albef}
\vspace{-0.4cm}
\end{table}

\subsection{Datasets}

\cref{tab:dataset_stats} lists statistics for each dataset in our study.
\vqavii~\cite{balanced_vqa_v2} is the most commonly used VQA dataset to date, it consists of 265K images and 1.1M question-image pairs, each with 10 ground-truth answers.
\vqacp~\cite{Agrawal_2018_CVPR} re-splits the \vqavii dataset such that, for every question type, train and test sets have different prior distributions of answers.
\vg~\cite{krishna2017visual} includes 108K images and 1.7M questions, each paired with a single answer, centered around either the full image or a specific region within it.
\gqa~\cite{Hudson_2019_CVPR} is another large-scale effort (22M questions, each with one answer) that focuses on compositionality of template-generated questions for real-world images (from \vg). Following prior work, we use the \gqa \emph{balanced} subset (1.5M questions).
Finally, \vizwiz~\cite{Gurari_2018_CVPR} is the only real-world VQA dataset as it was collected from visually impaired people. It consists of 31K image-question pairs, each paired with 10 answers.

\subsection{Training Details}

Following common practice, for discriminative models, we select the top-$k$ most frequent answers from the fine-tuning dataset, as the set of answer classes to perform classification over. Here $k$ is a dataset-dependent variable.
For \vqavii and \gqa, we use the same answer sets as \vilbert (3{,}129 and 1{,}533, respectively).
For \vizwiz, we select the answers that appear at least 8 times in training and validation sets, for a total of 3{,}112 answers that cover 97\% of the data.
For \vg, we select the answers that appear at least 29 times in the dataset, for a total of 3{,}449 answers that cover 76.5\% of the data.
Importantly, combined with the VQA accuracy metric defined above, this results in an upperbound to the accuracy that discriminative models can achieve in each dataset (see \cref{tab:max_score_overall_vilbert}).

All models are trained on the respective training sets and evaluated on the validation sets, which lets us conduct in-depth analyses that would otherwise be impossible to carry out on private test sets.
As there is no official split of \vg, we randomly sample the data into training (60\%) and validation (40\%) such that no image appears in both splits.

\section{Additional Results}
\label{app:more_results}

\paragraph{Evaluation with Shared Answer Sets}
\label{app:shared_answer_sets}
While different answer sets are an apparent issue for discriminative models, they also impact the performance of generative models, as the number of data points for each answer class seen by the generative model during fine-tuning varies: data-points in \topk{} answer set are more frequent than others (by definition of \topk{}).
In other words, even though a tokenizer used to produce an answer could generate it, it is unlikely (or less likely) to do so if it has not seen (or seen rarely) that combination of tokens during fine-tuning. Thus, even for generative models, we consider performance on \topk{} most frequent classes for each benchmark.

Thus, we report the accuracy on the subset of test questions whose answers are shared between \emph{both} the IID and the OOD models. 
For instance, when comparing the performance of the \vqavii and \vg fine-tuned models on the \vqavii test set, we compute the average accuracy on those \vqavii questions whose ground truth answers are present in the \topk{} answers from \vqavii as well as the \topk{} answers from \vg: we extract the common answer labels (between \vqavii and \vg \topk{} answers) and compute performance on test questions belonging to these shared answer labels only.

For IID evaluations, there are several possible ways to define shared answer sets based on OOD vocabs. While a subset is shown in \cref{fig:in_vocab}, \cref{tab:in_vocab_all} lists the VQA accuracy of each model in the IID settings when evaluated on the questions in the test sets whose answers are shared between the \topk{} answers in both the IID and the OOD settings (see \cref{subsec:shared_ans_set_eval} for more details).

In some IID cases, restricting the answer set to common answers hurts the performance (indicated as a lack of dotted bar in \cref{fig:in_vocab}). Interestingly, this pattern is observed across all models for some IID evaluations where the shared answer set is computed with respect to the \vg benchmark only: \vqavii for \dvilbert (-7.65 pp drop), \vqavii (-8.65 pp drop) and \gqa (-8.00 pp drop) for \gvilbert, \vqavii (-6.86 pp drop) for \dalbef, and \vqavii (-6.89 pp drop) for \galbef.
This seems to indicate that the \gqa and \vqavii questions corresponding to shared ans set with \vg are more difficult than the average difficulty of these test sets.

\begin{table}[t]
\vspace{-2pt}
\small
\setlength{\tabcolsep}{3.5pt}
\begin{subtable}[t]{0.48\textwidth}
    {\small
    \begin{center}
            \begin{tabular}{@{}l@{\hspace{4\tabcolsep}}rrrr}
            \toprule
             & \vqavii & \gqa & \vg & \vizwiz \\
             \midrule
            \vqavii & -- & 0.41 & 0.51 & 0.11$^\wedge$ \\
            \gqa & 0.25 & -- & 0.44 & 0.14$^\wedge$ \\
            \vg & 0.28 & 0.38 & -- & 0.03$^\wedge$ \\
            \vizwiz & 0.46 & 0.54 & 0.48 & -- \\
            \bottomrule
            \end{tabular}
    \end{center}
    }
    \vspace{-6pt}
    \caption{Discriminative \albef.}
    \vspace{0.3cm}
\end{subtable}
\begin{subtable}[t]{0.48\textwidth}
    {\small
    \begin{center}
            \begin{tabular}{@{}l@{\hspace{4\tabcolsep}}rrrr}
            \toprule
             & \vqavii & \gqa & \vg & \vizwiz \\
             \midrule
            \vqavii & -- & 0.45 & 0.48 & 0.26 \\
            \gqa & 0.26 & -- & 0.45 & 0.22 \\
            \vg & 0.27 & 0.35 & -- & 0.09$^\wedge$ \\
            \vizwiz & 0.48 & 0.56 & 0.52 & -- \\
            \bottomrule
            \end{tabular}
    \end{center}
    }
    \vspace{-6pt}
    \caption{Generative \vilbert.}
    \vspace{0.3cm}
\end{subtable}
\begin{subtable}[t]{0.48\textwidth}
    {\small
    \begin{center}
            \begin{tabular}{@{}l@{\hspace{4\tabcolsep}}rrrr}
            \toprule
             & \vqavii & \gqa & \vg & \vizwiz \\
             \midrule
            \vqavii & -- & 0.49 & 0.50 & 0.27 \\
            \gqa & 0.30 & -- & 0.45 & 0.34 \\
            \vg & 0.25 & 0.38 & -- & 0.16 \\
            \vizwiz & 0.50 & 0.57 & 0.51 & -- \\
            \bottomrule
            \end{tabular}
    \end{center}
    }
    \vspace{-6pt}
    \caption{Discriminative \vilbert.}
\end{subtable}
\caption{Spearman's rank correlation between drops in test accuracy (from IID to OOD) and the differences in proportion of answer classes between IID and OOD fine-tune sets. Unless otherwise specified with a $^\wedge$ character, $\rho$ values are significant with $p<.05$. Rows correspond to the fine-tuning datasets, columns correspond to the test benchmarks.} \label{tab:correlations_all}
\vspace{-0.4cm}
\end{table}

\paragraph{Answer Frequency Correlation} 
\label{app:ans_freq_corr}
\begin{table}[t!]
\setlength{\tabcolsep}{4.5pt}
{\small
\begin{center}

\begin{tabular}{@{}l@{\hspace{4\tabcolsep}}cccc}
\toprule
 & \vqavii & \gqa & \vg & \vizwiz \\
\midrule
\vqavii & -- & 0.43 & 0.51 & 0.25 \\
\gqa & 0.27 & -- & 0.43 & 0.19 \\
\vg & 0.26 & 0.36 & -- & 0.13 \\
\vizwiz & 0.47 & 0.55 & 0.48 & -- \\
\bottomrule
\end{tabular}

\end{center}
}
\caption{Spearman's rank correlation between drops in test accuracy (from IID to OOD) and the differences in proportion of answer classes between IID and OOD fine-tuning sets for \galbef. $p<.05$ for all $\rho$. Rows correspond to the fine-tuning datasets, columns correspond to the test benchmarks.} \label{tab:acc_freq_corr_albef_gen}
\vspace{-0.4cm}
\end{table}

In order to examine the relationship between accuracy drop for less frequent classes, we first compute per answer-class accuracy (average accuracy of all test questions belonging to the same answer class) for answers in shared answer set. 
We then sort the shared answer classes based on their weighted drop in per-class accuracy from IID to OOD (IID accuracy - OOD accuracy), \ie{} absolute drop in per-class accuracy weighted by number of data points belonging to that class in the test set.
We then compute the Spearman's rank correlation of these weighted drop in per-class accuracies with difference in percentage frequencies of the answer classes between IID and OOD fine-tuning sets (percentage frequency of an answer class in IID minus its percentage frequency in OOD).

\cref{tab:correlations_all} list Spearman's rank correlations of IID-to-OOD drops in test accuracy vs. proportion of answer classes in respective (IID and OOD) fine-tuning sets for \dalbef, \gvilbert{} and \dvilbert{} (see \cref{subsec:shared_ans_set_eval} for more details).
As a simple baseline test, we also compute correlations and p-values for a permuted dataset to confirm their lack of significance, or correlation values close to zero.
\paragraph{Maximum Achievable Scores}

\begin{figure*}
\centering
    \includegraphics[width=\textwidth]{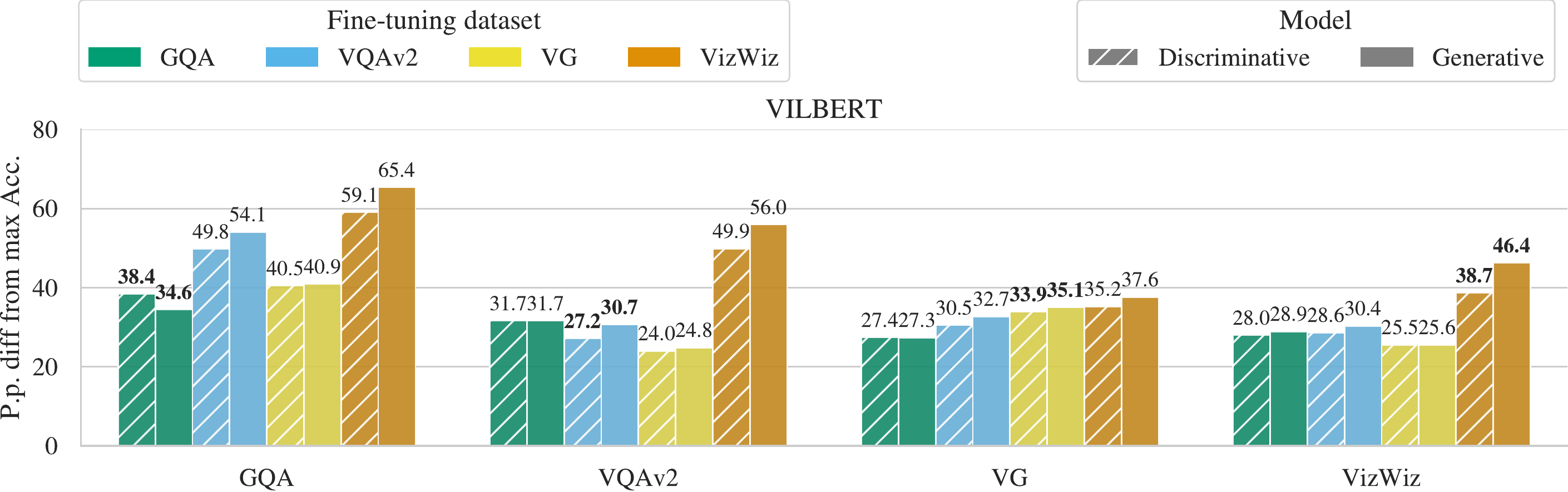}
    \includegraphics[width=\textwidth]{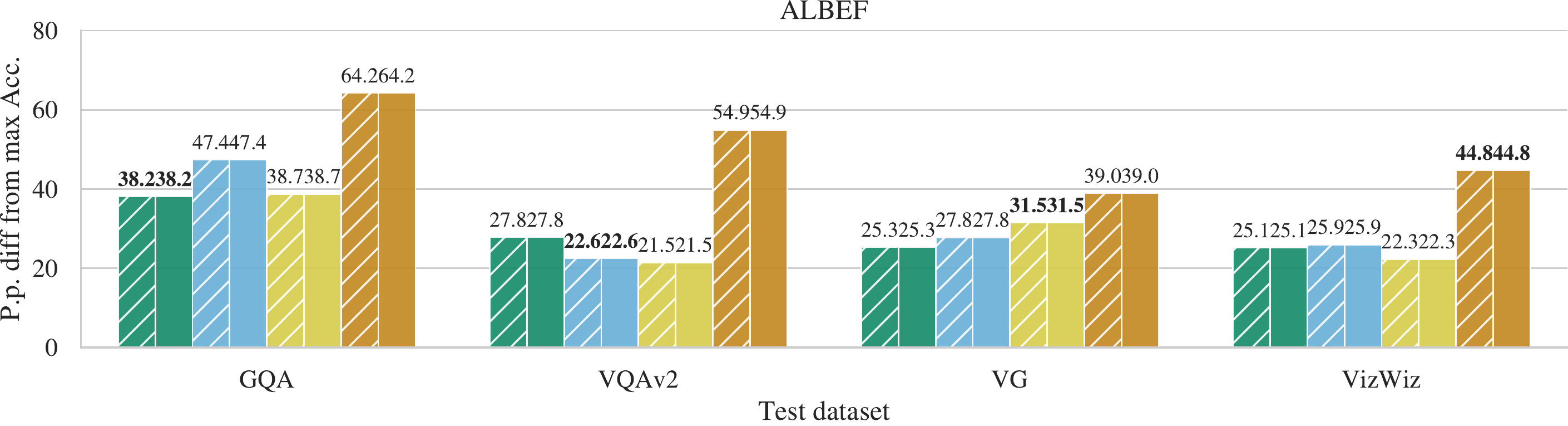}    

\caption{
Percentage point difference between maximum achievable accuracies in Table~\ref{tab:max_score_overall_vilbert} and accuracies in Figure~\ref{fig:iid_ood}. Results for \vilbert pretrained on the same data as \albef 4M are also shown.}
\label{fig:diff_iid_ood}
\end{figure*}

Table~\ref{tab:max_score_overall_vilbert} lists maximum achievable accuracies, and Figure~\ref{fig:diff_iid_ood} shows the difference between those scores and bar values shown in Figure~\ref{fig:iid_ood}.
In our analyses, we also noted that differences in answer pre-processing strategies can result in slightly different numbers than those reported in \cref{tab:max_score_overall_vilbert}. However, those differences did not change the conclusion of our findings.

\begin{figure*}[t]
\centering
    \includegraphics[width=\textwidth]{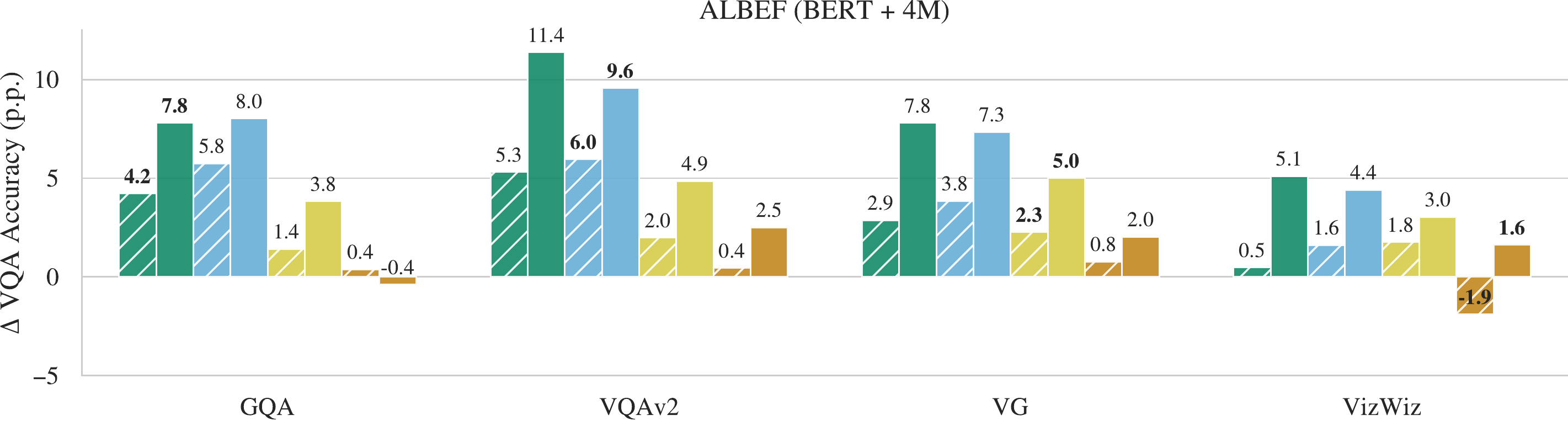}
\caption{
Difference in VQA accuracy (p.p.) for \albef that has and has not been pretrained on the 4M dataset.}
\label{fig:pretraining_effect_albef_small}
\end{figure*}

\paragraph{Effect of pretraining data size on \albef}
For the \albef model, while we often observe improvements by increasing the size of the multimodal pretraining dataset (4M vs. 14M), the improvements are small. 
When pretraining on the smaller dataset (4M, see \cref{fig:pretraining_effect_albef_small}), we observe a median improvement (over no pretraining) of ~1.9\% for the discriminative and ~4.9\% for the generative \albef, while the median additional improvements due to larger pretraining dataset (14M) are ~0.1\% and ~0.6\% respectively (refer to \cref{fig:pretraining_effect}).
Surprisingly, there are also dataset pairs for which larger pretraining has a negative effect when compared to the performance with a smaller pretraining set (\eg, \albef model fine-tuned on \vizwiz and tested on \vqavii).

\begin{table*}[t!]
    \setlength{\tabcolsep}{2.5pt}
    \centering
    \small
    \begin{tabular}{llllccc}
\toprule
          Model &     Test & Fine-tune &              Answer Set &  VQA Acc. (IID) &  VQA Acc. (OOD) & Difference \\
\midrule
 \band\dvilbert &     \gqa &      \gqa &     \gqa $\cap$ \vqavii &           63.05 &           48.53 &      14.52 \\
      \dvilbert &     \gqa &      \gqa &         \gqa $\cap$ \vg &           52.32 &           35.08 &      17.24 \\
      \dvilbert &     \gqa &      \gqa &     \gqa $\cap$ \vizwiz &           64.91 &           28.39 &      36.52 \\
 \band\dvilbert &      \vg &       \vg &     \vg $\cap$ \vqavii &           58.18 &           51.97 &       6.21 \\
      \dvilbert &      \vg &       \vg &         \vg $\cap$ \gqa &           59.57 &           39.15 &      20.42 \\
      \dvilbert &      \vg &       \vg &      \vg $\cap$ \vizwiz &           60.65 &           13.23 &      47.42 \\
 \band\dvilbert &  \vizwiz &   \vizwiz &  \vizwiz $\cap$ \vqavii &           51.78 &           22.37 &      29.41 \\
      \dvilbert &  \vizwiz &   \vizwiz &     \vizwiz $\cap$ \gqa &           51.05 &           15.40 &      35.65 \\
      \dvilbert &  \vizwiz &   \vizwiz &      \vizwiz $\cap$ \vg &           50.07 &           13.59 &      36.48 \\
      \dvilbert &  \vqavii &   \vqavii &     \vqavii $\cap$ \gqa &           71.74 &           52.42 &      19.32 \\
 \band\dvilbert &  \vqavii &   \vqavii &      \vqavii $\cap$ \vg &           58.10 &           48.84 &       9.26 \\
      \dvilbert &  \vqavii &   \vqavii &  \vqavii $\cap$ \vizwiz &           68.20 &           33.87 &      34.33 \\
      \midrule
      \gvilbert &     \gqa &      \gqa &     \gqa $\cap$ \vqavii &           67.01 &           44.03 &      22.98 \\
 \band\gvilbert &     \gqa &      \gqa &         \gqa $\cap$ \vg &           57.35 &           34.40 &      22.95 \\
      \gvilbert &     \gqa &      \gqa &     \gqa $\cap$ \vizwiz &           69.63 &           20.73 &      48.90 \\
 \band\gvilbert &      \vg &       \vg &      \vg $\cap$ \vqavii &           55.52 &           47.93 &       7.59 \\
      \gvilbert &      \vg &       \vg &         \vg $\cap$ \gqa &           57.65 &           39.26 &      18.39 \\
      \gvilbert &      \vg &       \vg &      \vg $\cap$ \vizwiz &           58.84 &            7.67 &      51.17 \\
 \band\gvilbert &  \vizwiz &   \vizwiz &  \vizwiz $\cap$ \vqavii &           43.06 &           19.58 &      23.48 \\
      \gvilbert &  \vizwiz &   \vizwiz &     \vizwiz $\cap$ \gqa &           42.57 &           13.71 &      28.86 \\
      \gvilbert &  \vizwiz &   \vizwiz &     \vizwiz $\cap$ \vg &           41.11 &           13.20 &      27.91 \\
      \gvilbert &  \vqavii &   \vqavii &     \vqavii $\cap$ \gqa &           68.01 &           52.35 &      15.66 \\
 \band\gvilbert &  \vqavii &   \vqavii &      \vqavii $\cap$ \vg &           53.59 &           46.78 &       6.81 \\
      \gvilbert &  \vqavii &   \vqavii &  \vqavii $\cap$ \vizwiz &           64.69 &           26.82 &      37.87 \\
      \midrule
   \band\dalbef &     \gqa &      \gqa &     \gqa $\cap$ \vqavii &           63.24 &           51.06 &      12.18 \\
        \dalbef &     \gqa &      \gqa &         \gqa $\cap$ \vg &           53.09 &           37.97 &      15.12 \\
        \dalbef &     \gqa &      \gqa &     \gqa $\cap$ \vizwiz &           64.85 &           22.14 &      42.71 \\
   \band\dalbef &      \vg &       \vg &       \vg $\cap$ \vqavii &           61.38 &           55.83 &       5.55 \\
        \dalbef &      \vg &       \vg &        \vg $\cap$ \gqa &           63.76 &           43.82 &      19.94 \\
        \dalbef &      \vg &       \vg &      \vg $\cap$ \vizwiz &           64.12 &            4.52 &      59.60 \\
   \band\dalbef &  \vizwiz &   \vizwiz &  \vizwiz $\cap$ \vqavii &           42.42 &           26.23 &      16.19 \\
        \dalbef &  \vizwiz &   \vizwiz &     \vizwiz $\cap$ \gqa &           40.49 &           20.44 &      20.05 \\
        \dalbef &  \vizwiz &   \vizwiz &      \vizwiz $\cap$ \vg &           38.15 &           19.68 &      18.47 \\
        \dalbef &  \vqavii &   \vqavii &     \vqavii $\cap$ \gqa &           76.64 &           57.18 &      19.46 \\
   \band\dalbef &  \vqavii &   \vqavii &      \vqavii $\cap$ \vg &           63.47 &           52.78 &      10.69 \\
        \dalbef &  \vqavii &   \vqavii &  \vqavii $\cap$ \vizwiz &           72.84 &           28.08 &      44.76 \\
        \midrule
   \band\galbef &     \gqa &      \gqa &      \gqa $\cap$ \vqavii &           65.81 &           51.72 &      14.09 \\
        \galbef &     \gqa &      \gqa &         \gqa $\cap$ \vg &           56.08 &           37.71 &      18.37 \\
        \galbef &     \gqa &      \gqa &     \gqa $\cap$ \vizwiz &           67.05 &           27.61 &      39.44 \\
   \band\galbef &      \vg &       \vg &      \vg $\cap$ \vqavii &           62.71 &           57.33 &       5.38 \\
        \galbef &      \vg &       \vg &         \vg $\cap$ \gqa &           65.48 &           51.17 &      14.31 \\
        \galbef &      \vg &       \vg &      \vg $\cap$ \vizwiz &           66.20 &           21.13 &      45.07 \\
   \band\galbef &  \vizwiz &   \vizwiz &  \vizwiz $\cap$ \vqavii &           52.85 &           28.96 &      23.89 \\
        \galbef &  \vizwiz &   \vizwiz &     \vizwiz $\cap$ \gqa &           52.58 &           22.21 &      30.37 \\
        \galbef &  \vizwiz &   \vizwiz &      \vizwiz $\cap$ \vg &           51.94 &           23.56 &      28.38 \\
        \galbef &  \vqavii &   \vqavii &     \vqavii $\cap$ \gqa &           78.03 &           62.93 &      15.10 \\
   \band\galbef &  \vqavii &   \vqavii &      \vqavii $\cap$ \vg &           65.20 &           55.64 &       9.56 \\
        \galbef &  \vqavii &   \vqavii &  \vqavii $\cap$ \vizwiz &           74.38 &           39.37 &      35.01 \\
\bottomrule
\end{tabular}
    \caption{VQA accuracy of each model in the IID settings (see column VQA Acc. (IID)) when evaluated on the questions in the test sets whose answers are shared between the \topk{} answers in both the IID and the OOD settings. Please refer to \cref{subsec:shared_ans_set_eval} for more details.
    Answer Set: OOD benchmarks with respect to which IID shared answer set accuracy is computed.
    VQA Acc. (OOD): OOD accuracy on questions corresponding to the shared answer set, \ie{} when fine-tuned on the OOD dataset mentioned in Answer Set column and tested on the benchmark mentioned in the Test column. 
    Difference: VQA Acc. (IID) - VQA Acc. (OOD). Gray bands highlight the OOD benchmarks with respect to which IID shared answer set accuracy is computed in \cref{fig:in_vocab}.}
    \label{tab:in_vocab_all}
\end{table*}

\section{Potential Causes of Poor OOD Generalization: A Qualitative Study}
\label{app:qual_study}

In section \ref{sec:ood_gen}, we observe that our pretrained models exhibit poor OOD generalization for the task of VQA. We also noted that this poor generalization is not entirely explained by the absence or poor representation of test answer classes in the training data. Here, we perform a qualitative study to dig deeper into the potential causes of the poor OOD generalization. We manually examine 20 randomly-sampled qualitative examples of failure cases on top-30 answer classes contributing the most to the drop in performance from IID to OOD. We only focus on answer classes that are shared between the train and test splits to make sure the performance drop is not due to the absence of answer classes in the training dataset. We report the top-5 classes that contribute the most to the drop in performance for each OOD setting in \cref{tab:top_ans_class_all}. Below, we describe four major potential causes\footnote{For poor OOD generalization on the \vizwiz benchmark, one of the reasons could be difference in image distributions between \vizwiz (that contains many blurry pictures, or pictures with poor lighting conditions) and other three datasets (that contain clear pictures).} for the poor OOD generalization that we can infer from our qualitative study on \dvilbert\footnote{We use the model trained with the official codebase.} 
and \galbef. The specific examples reported below are for \dvilbert.

\paragraph{Poor reasoning skills.} In \cref{tab:top_ans_class_all}, we can see  that a model fine-tuned on \vqavii, \vg, or \vizwiz and evaluated on \gqa shows the highest performance drop on classes such as ``yes'', ``no'', ``right'', ``left'', ``top'', and ``bottom''. For instance, \vqavii--\gqa (fine-tuned on \vqavii, evaluated on \gqa) model underperforms \gqa-\gqa model by 24\% for ``no.'' Upon qualitative examination, we find that for many of such failure cases, the \gqa questions are more compositional and hence require more complex reasoning (\eg, ``Are there both bison and zebras in the image?'', ``Is the cheese to the right or to the left of the empty plate?'') than the questions for the same answer classes in other datasets (\eg, from \vqavii train set: ``Is the TV turned on?'', ``Which hand is the man holding up?''). 
This study re-affirms previous findings \cite{Johnson_2017_CVPR,Hudson_2019_CVPR} -- VQA models lack sufficient logical, spatial, and compositional reasoning skills -- for the more recent, pretrained Transformer models.

\paragraph{Overfitting to the answer priors.} Previous studies have shown that VQA models tend to be biased towards the prior distribution of answers in the training set (per question type) \cite{Agrawal_2018_CVPR}. We find that this limitation exists in the more recent pretrained models as well, and it is especially hurtful in the OOD settings because the priors need not be the same across train and test sets, unlike in the IID settings. For instance, \dvilbert fine-tuned on \vqavii predicts ``2'' for a lot questions with target answer ``1'' in the \vg test set. Similarly, sometimes \dvilbert fine-tuned on \vg incorrectly predicts ``helmet'' for \vqavii test questions such as ``What is the skateboarder wearing to protect his head?'', ``What protective gear is he wearing?'' when the skateboarder is not wearing anything. This indicates that the model is relying on answer priors rather than visual grounding. Our experimental results on VQA-CP (\cref{sec:vqacp}) directly quantify the extent of such limitations in current models. 

\paragraph{Overfitting to the question format.} For each answer class, there is usually a limited variation in the format of questions in the fine-tuning set. For some of the answer classes showing poor OOD generalization, we found that certain question formats are quite dominant in the fine-tuning set, and that these dominant formats are different between the OOD fine-tuning and test sets.
Thus, we conjecture that models are likely overfitting to such dominant formats in fine-tuning data and hence fail to generalize at test time when the format changes. 
For instance, questions about ``chair'' in the \vqavii fine-tuning set are mostly of the form ``What is \dots{} sitting on''? whereas in the \gqa test set, they are mostly of the form ``What kind of furniture is \dots?''. Thus, the ``chair'' class accuracy of \dvilbert fine-tuned on \vqavii drops from 48\% when tested on \vqavii to 38\% on the \gqa test set. As another example, \dvilbert trained on \gqa fails terribly for ``dog'' and ``cat'' classes on \vg test set (accuracy drops of 47\% and 43\% respectively, where drop is between \gqa--\gqa and \gqa--\vg). \gqa questions are mostly of the form ``What animal \dots?'' or ``What kind of animal \dots?'' whereas \vg questions often do not mention the word ``animal'' and are of the form ``Who is \dots?'' or ``What is \dots?'' (\eg, ``Who is holding the Frisbee?'', ``What is on the leash?''). Similarly, for the answer class ``pizza'', \dvilbert fine-tuned on \vg has mostly seen questions of the format ``What food is this?'', ``What is the man eating?'', ``What is on the plate?'', ``What’s in the box?'' in \vg fine-tuning set. However, when evaluated on the \vqavii test set, the model fails to respond correctly for questions about ``pizza'' such as, ``What snack is this?'' (model response: ``pineapple''), ``What recipe this will become?'' (model response: ``cheese''), ``What’s in the bowl'' (model response: ``tomato sauce''). For the last example, perhaps the model is not expecting pizza to be in a bowl. 

Related to above, we observed that sometimes \dvilbert fails to produce the correct answer type for a given question. For instance, \dvilbert fine-tuned on \vg responds with ``woman'' to the question ``Is the person who is cutting these carrots right handed or left handed?''. So it appears as if the \vg model does not understand the question structure in this example, \ie{} the response is expected to be either ``right'' or  ``left''. Similarly, for the question ``Are there more blue or black shirts?'', \vg model responds with ``rolled up''. Similarly, it answers ``1 on right'' to the question ``What type of apple is shown?'', instead of describing some attribute of apple such as ``green''.

\paragraph{Stringent evaluation metric.} We notice that sometimes the models' responses are correct but they are evaluated as incorrect because those responses do not exist in the ground-truth answers. For instance, \vqavii--\vg model gets penalized for answering ``table'' instead of ``on table''\footnote{Note that before computing the accuracy, both the predicted and the ground truth answers are pre-processed for answer normalization but such pre-processing is very basic. More details of the pre-processing can be found at \url{https://visualqa.org/evaluation.html}} (Q: ``Where is \dots ?'') or ``sunny'' instead of ``clear'' (Q: ``How is the weather?''). More examples in \cref{fig:strict_eval_metric}. This effect is expected to be more pronounced for the OOD evaluation than IID, because in IID a model can learn the format of the test answer (``on table'' vs. ``table'', ``clear vs. sunny'') from the train set, whereas in OOD the format in the train set can be different from the test set. Also, such stringent evaluation (\ie, performing string matching with a small set of ground-truth answers) is expected to hurt generative models more than discriminative ones because they show more variations in the form of the answers as they are not limited by a fixed answer vocabulary (\eg., ``pizza slices'' instead of ``pizza'' (Q: ``What are these?''), ``pizzeria'' instead of ``pizza'' (Q: ``What kind of restaurant is this?''). We observed that, \vg model (model fine-tuned on \vg) evaluated on \gqa answers questions about ``man'' with ``snowboarder'', ``man on left'' (i.e. more descriptive referring expressions) than just saying ``man'' but it does not get any credit because \gqa ground truth is ``man''. To quantify the extent of this issue and measure its effect on discriminative vs. generative models, IID vs. OOD settings, we perform human evaluation of machine generated answers and provide additional insights in \cref{sec:human_eval}.  

\paragraph{Poor performance of GQA model on color questions (both IID and OOD):} \dvilbert fine-tuned on \gqa does not seem to be transferring well to color questions in the \vqavii and \vg test sets (and even in IID \gqa test set). In \cref{tab:top_ans_class_all}, we can see that the top-5 answer classes with highest drop in IID-to-OOD performance for \gqa model have quite a few colors. For instance, for the answer class ``red'' in the \vg test set, \gqa model fails to correctly answer simple questions (given the kind of questions \gqa model is fine-tuned on) such as ``What is the primary color of the sign on the right?'', ``What is the main color of the strawberry?'', ``What color is the pull luggage of the woman?'', ``What color are the pepperonis?''. It is not clear why \gqa model does not perform well on color questions.

\begin{table*}[]
\center
\resizebox{0.8\textwidth}{!}{
\begin{tabular}{llll}
\toprule
\textbf{Train data}   & \textbf{Test data}      & \textbf{Model}                    & \textbf{Answer classes}   \\
\midrule
\multirow{15}{*}{VQAv2} & \multirow{5}{*}{GQA}  & Discriminative ViLBERT            & no, yes, left, right, top \\
                      &                         & Discriminative ViLBERT (in-house) & no, yes, right, bottom, color \\
                      &                         & Generative ViLBERT (in-house)     & no, yes, right, left, bottom \\
                      &                         & Discriminative ALBEF              & no, left, yes, bottom, chair \\
                      &                         & Generative ALBEF                  & left, no, yes, bottom, top \\
                      \cline{2-4}
                      & \multirow{5}{*}{VG}     & Discriminative ViLBERT            & 1, no 1, daytime, on table, in sky \\
                      &                         & Discriminative ViLBERT (in-house) & daytime, 1, white, 2, black \\
                      &                         & Generative ViLBERT (in-house)     & daytime, white, 2, black, 1 \\
                      &                         & Discriminative ALBEF              & 1, daytime, in sky, on table, white \\
                      &                         & Generative ALBEF                  & 1, daytime, black, in sky, clear \\
                      \cline{2-4}
                      & \multirow{5}{*}{VizWiz} & Discriminative ViLBERT            & no, blue, yes, white, black \\
                      &                         & Discriminative ViLBERT (in-house) & yes, black, water bottle, corn, soup \\
                      &                         & Generative ViLBERT (in-house)     & pink, brown, corn, wine, keys \\
                      &                         & Discriminative ALBEF              & keyboard, no, soup, cake, samsung \\
                      &                         & Generative ALBEF                  & soup, lotion, black, brown, corn \\
\midrule
\multirow{15}{*}{GQA} & \multirow{5}{*}{VQAv2}  & Discriminative ViLBERT            & yes, no, white, red, black \\
                      &                         & Discriminative ViLBERT (in-house) & no, yes, white, red, tennis \\
                      &                         & Generative ViLBERT (in-house)     & no, yes, white, red, tennis \\
                      &                         & Discriminative ALBEF              & no, yes, white, red, right \\
                      &                         & Generative ALBEF                  & no, yes, right, red, black and white \\
                      \cline{2-4}
                      & \multirow{5}{*}{VG}     & Discriminative ViLBERT            & white, trees, green, black, black and white \\
                      &                         & Discriminative ViLBERT (in-house) & white, black, trees, green, blue \\
                      &                         & Generative ViLBERT (in-house)     & white, trees, black, green, brown \\
                      &                         & Discriminative ALBEF              & white, trees, black and white, grass, green \\
                      &                         & Generative ALBEF                  & trees, green, black and white, black, grass \\
                      \cline{2-4}
                      & \multirow{5}{*}{VizWiz} & Discriminative ViLBERT            & no, blue, yes, white, laptop \\
                      &                         & Discriminative ViLBERT (in-house) & blue, white, black, dog, laptop \\
                      &                         & Generative ViLBERT (in-house)     & white, blue, laptop, black, dog \\
                      &                         & Discriminative ALBEF              & no, keyboard, soup, red, cake \\
                      &                         & Generative ALBEF                  & no, dog, keyboard, laptop, blue \\
\midrule
\multirow{15}{*}{VG} & \multirow{5}{*}{VQAv2}   & Discriminative ViLBERT            & 0, white, nothing, gray, red \\
                      &                         & Discriminative ViLBERT (in-house) & 0, 3, left, nothing, brown \\
                      &                         & Generative ViLBERT (in-house)     & 0, 1, gray, left, wii \\
                      &                         & Discriminative ALBEF              & 0, nothing, left, brown, 2 \\
                      &                         & Generative ALBEF                  & 0, 3, nothing, right, gray \\
                      \cline{2-4}
                      & \multirow{5}{*}{GQA}    & Discriminative ViLBERT            & right, left, bottom, top, gray \\
                      &                         & Discriminative ViLBERT (in-house) & bottom, left, top, color, large \\
                      &                         & Generative ViLBERT (in-house)     & left, bottom, color, top, gray \\
                      &                         & Discriminative ALBEF              & left, bottom, top, black, chair \\
                      &                         & Generative ALBEF                  & left, bottom, color, top, gray \\
                      \cline{2-4}
                      & \multirow{5}{*}{VizWiz} & Discriminative ViLBERT            & blue, black, grey, red, soup \\
                      &                         & Discriminative ViLBERT (in-house) & grey, black, blue, white, computer screen \\
                      &                         & Generative ViLBERT (in-house)     & grey, blue, black, pink, computer screen \\
                      &                         & Discriminative ALBEF              & grey, soup, remote, cake, samsung \\
                      &                         & Generative ALBEF                  & grey, blue, soup, wine, pink \\
\midrule
\multirow{15}{*}{VizWiz} & \multirow{5}{*}{VQAv2} & Discriminative ViLBERT          & no, yes, 1, 2, white \\
                      &                         & Discriminative ViLBERT (in-house) & no, 1, 2, 0, white \\
                      &                         & Generative ViLBERT (in-house)     & no, yes, 1, 2, white \\
                      &                         & Discriminative ALBEF              & no, 1, 2, yes, blue \\
                      &                         & Generative ALBEF                  & yes, no, 1, 2, 0 \\
                      \cline{2-4}
                      & \multirow{5}{*}{GQA}    & Discriminative ViLBERT            & no, right, left, man, bottom \\
                      &                         & Discriminative ViLBERT (in-house) & no, right, bottom, man, top \\
                      &                         & Generative ViLBERT (in-house)     & no, yes, left, right, man \\
                      &                         & Discriminative ALBEF              & no, left, bottom, top, man \\
                      &                         & Generative ALBEF                  & yes, left, no, bottom, top \\
                      \cline{2-4}
                      & \multirow{5}{*}{VG}     & Discriminative ViLBERT            & 1, white, green, 2, black \\
                      &                         & Discriminative ViLBERT (in-house) & 1, 2, white, green, man \\
                      &                         & Generative ViLBERT (in-house)     & 1, 2, white, green, black \\
                      &                         & Discriminative ALBEF              & 1, green, white, 1, 2, blue \\
                      &                         & Generative ALBEF                  & 1, 2, black, man, green \\
\bottomrule
\end{tabular}
}
\caption{Top-5 answer classes with highest performance drop from IID to OOD (for the same test set) for all OOD configurations. The answer classes are sorted by drop in weighted (wtd) accuracy, i.e. drop in absolute (abs) accuracy weighted by the \# test questions for that answer class.}
\label{tab:top_ans_class_all}
\end{table*}

\section{Human Evaluation}
\label{app:human_eval}

\paragraph{Method.}
We used Amazon Mechanical Turk to collect human judgment about model responses on a random subset of 10K questions for each of the test sets---\vqavii, \gqa and \vg. Since the size of \vizwiz test set is less than 10K, we collected human judgment on all the \vizwiz test questions. However, we dropped the questions that were tagged as ``unanswerable'' or ``unsuitable'' (more details are provided below under ``Filtering VizWiz data''). The total number of \vizwiz test questions for which we collected human judgment is 1440 (per model). We performed human evaluation of the responses from the following models -- \dvilbert\footnote{For \dvilbert, we had initially collected human judgements for the version trained using the official codebase, and we did not collect annotations again for our re-implementation due to time constraints. Given our results above, we do not expect significant differences between the two versions.}  and \gvilbert trained on the \vqavii, \gqa, \vg datasets.
We did not collect human judgments for models \emph{fine-tuned} on \vizwiz, because a significant proportion of the responses from these models tend to be ``unanswerable'' or ``unsuitable'' (35\% on \vqavii, 39\% on \gqa, 65\% on \vg, and 64\% on \vizwiz). 
Collecting human feedback about such responses would not provide useful insights, because all questions in \vqavii, \gqa and \vg should be answerable, therefore all cases of ``unanswerable'' should be incorrect. Such responses are just a side effect of a model's priors caused by all the unanswerable training points in the \vizwiz fine-tuning set.

For each response, we asked 5 raters to evaluate the question, image, and a given model response, and indicate through a binary choice whether they considered the model response a correct answer to the question or not. To control the quality of the data, we filtered out low quality data using different heuristics such as distribution of yes/no answers for each worker, their mean submission times, average agreement with their fellow workers, or average alignment with the automatic accuracy.\footnote{How frequently a worker's response (yes/no) aligns with the automatic accuracy computed (100.0/0.0) More specifically, we equate the worker's \emph{yes} response with 100.0 and \emph{no} with 0.0 and look at the average difference between the worker's response and the automatic accuracy}. In each of these cases, we looked at random samples from the outliers to qualitatively confirm our hypothesis. More details about the human evaluation interface are presented in the next paragraph.

To compute human accuracy of a model response (for a given question and image), we considered a response correct if at least 4 raters voted it is correct, and incorrect otherwise. We decided so in order to decrease noise introduced by cases where there was low agreement between raters. 

\begin{figure*}[t!]
\centering
    \includegraphics[width=\textwidth, trim={0cm 0.7cm 0cm 0cm}, clip]{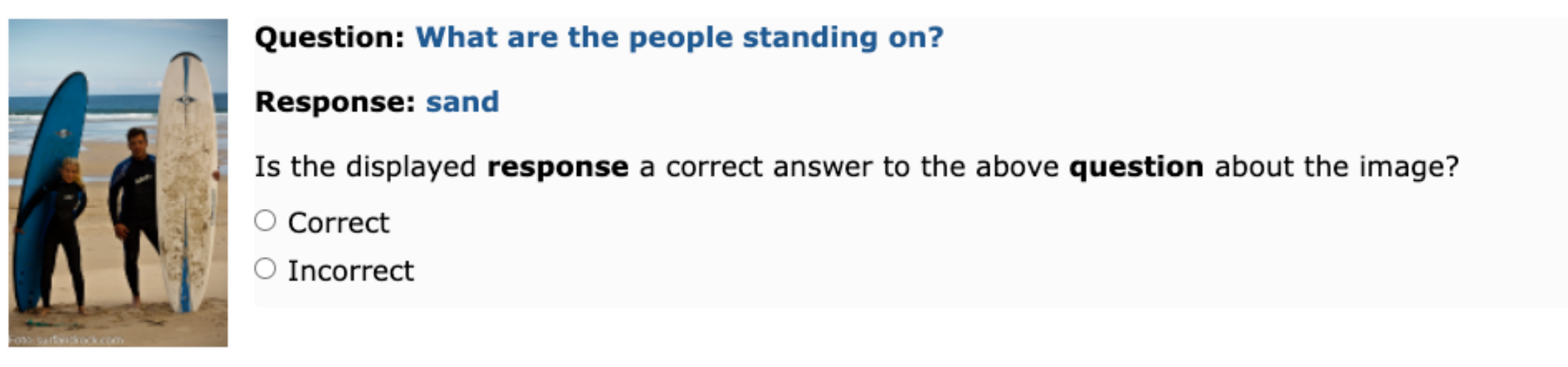}
    \caption{Sample of the MTurk interface the raters used to annotate data.} \label{fig:mturk_interface}
    \vspace{0.3cm}
    \includegraphics[width=0.94\textwidth, trim={0cm 2.5cm 0cm 0cm}, clip]{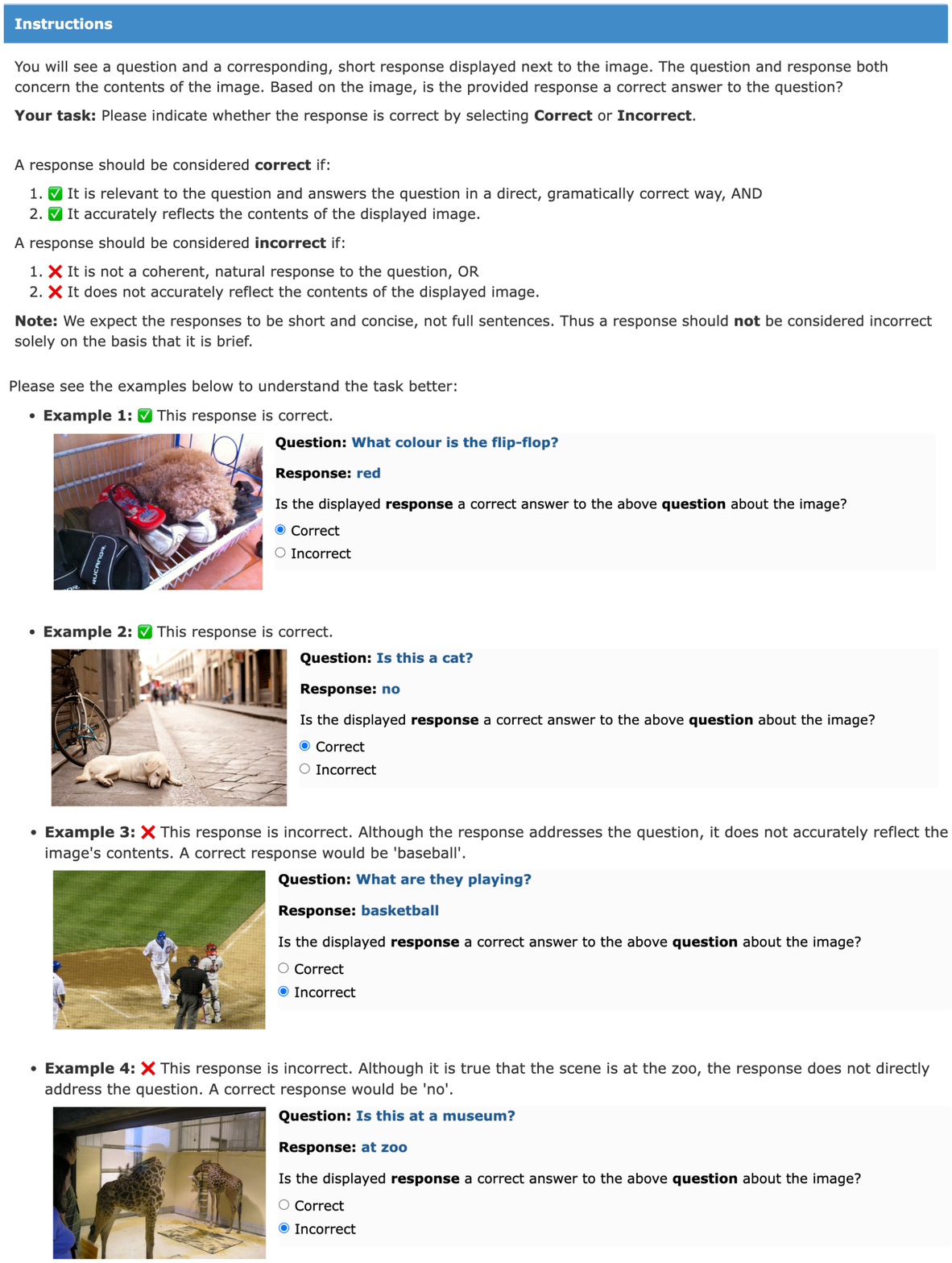}
    \vspace{-0.1cm}
    \caption{Instructions given to MTurk raters.} \label{fig:mturk_instructions}
\end{figure*}

\paragraph{Data collection interface.}
\cref{fig:mturk_interface} shows a sample of the interface the MTurk raters used to submit their responses.
The workers were shown some examples, but in order not to bias them, we did not give them detailed guidance as to what should be considered correct for not - rather we asked them to rely on common sense, and consider an answer correct if it seems both factually accurate and natural in the context. See \cref{fig:mturk_instructions} for details.

\paragraph{Filtering VizWiz data.}
For human evaluation we filtered out all image-question pairs for which the ground truth answer indicates it is \emph{unanswerable}. That is, we have not collected human feedback for questions for which the ground truth answer appears in the following list:

\begin{itemize}[noitemsep, topsep=2.5pt]
    \item "unanswerable", "unsuitable"
    \item "insufficient image"
    \item "unknown", "unsure", "not clear"
    \item "blurry", "too blurry"
    \item  "i don't know", "don't know", "i don no", "no idea"
    \item "unusable image",  "unsuitable image", "unstable image",   "insufficient image quality", "unreadable"
    \item  "i can't tell", "can't tell", "can't see"
    \item "0" \footnote{Qualitative examples have suggested that often this was used to indicate \emph{unanswerable}.}
\end{itemize}

In particular, this left us with 1440 questions for the \vizwiz dataset.

\begin{figure*}[t]
\centering
    \includegraphics[width=0.9\textwidth]{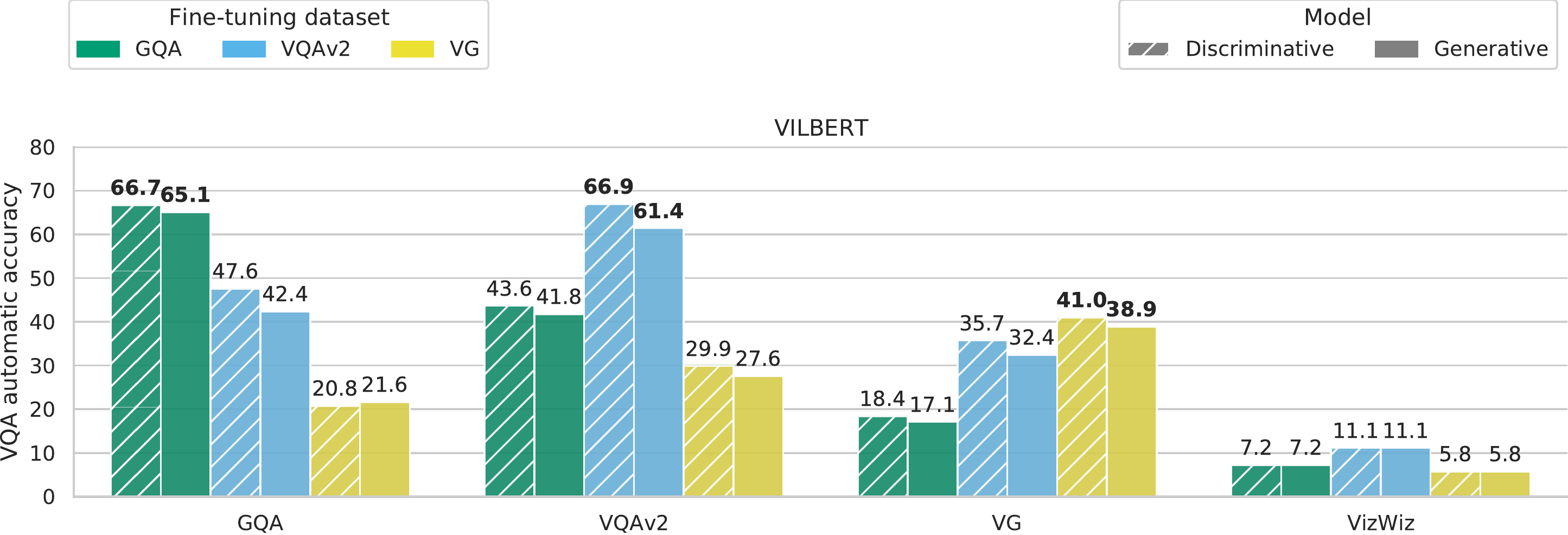}
    \includegraphics[width=0.9\textwidth]{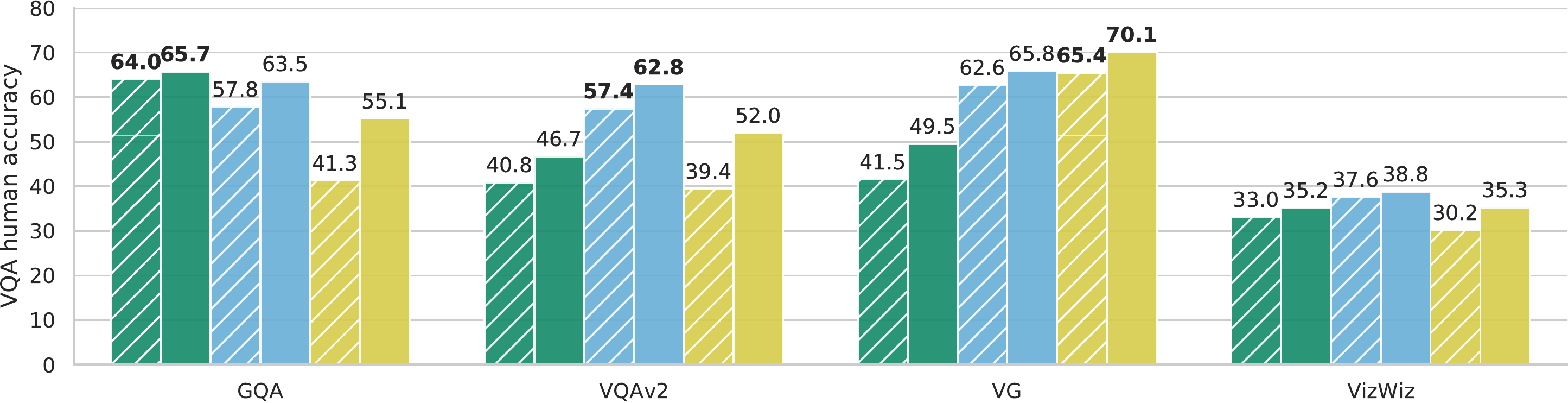}
    \caption{Human evaluation of \dvilbert (shaded bars) and \gvilbert (plain bars) and comparison with automatic evaluation on a random subset (10K) of test questions for each test dataset -- \gqa, \vqavii, \vg and \vizwiz. Accuracies in bold denote the IID settings. Top: accuracies obtained using automatic evaluation. Bottom: accuracies obtained using human evaluation.}
    \label{fig:human_results}
    \vspace{-0.3cm}
\end{figure*}

\paragraph{Results.}
We report the human accuracies for \dvilbert and \gvilbert in \cref{fig:human_results} (bottom). We also report the accuracies obtained using automatic metrics (please see \cref{subsec:dataset_eval_metric} for description of automatic metrics for each dataset) computed over the same random subset of test questions as that used for human evaluation in \cref{fig:human_results} (top).
Please refer to \cref{sec:human_eval} for discussion of results.

\paragraph{Qualitative examples of questions being incorrectly penalized by automatic evaluation}
\cref{table:5_yes} shows some examples for responses which were awarded $0.0$ accuracy using automatic metrics but were marked as correct by all 5 raters during human evaluation.

\paragraph{Discussion on VQA data quality}
For the collected human judgement data, we find that for a significant number of questions (32\%) there was low agreement between the 5 raters, i.e. either 3/5 answered \emph{correct} while the remaining 2/5 answered \emph{incorrect}, or vice-versa. Note that this is after we already filtered out low quality judgements. We have to recognize that, despite our best efforts to control data quality using our heuristics, there might still be low quality data in our dataset. Low quality annotations can be misleading and might distort the results of our analysis. Yet, we believe that we have collected a large enough sample to dampen the effect of these on the reported results.
Upon examining some examples from such low agreement questions, we find that many such cases highlight the quality of the VQA data. For instance, questions not being sufficiently objective but up for interpretation, questions phrased poorly that make it difficult to understand what the question is asking about, \etc{}. We discuss these further below.

\begin{itemize}[noitemsep, topsep=2.5pt]
    \item \textbf{Low agreement due to ambiguity.}
One reason why human raters might give different feedback stems from ambiguity and subjectivity around the question and the contents of the image. In these cases it is up to the raters subjective opinion to judge whether the answer is acceptable or not. Find some qualitative examples in \cref{table:ambiguous}.
    \item \textbf{Low agreement due to poor quality question.}
Some questions in the original dataset are of rather poor quality which makes it near impossible for the rater to provide a valuable response. Find some qualitative examples of such questions in \cref{table:poor}.
\end{itemize}

Also surprisingly, from \cref{fig:human_results_subset}, we see that for models fine-tuned on \vqavii or \gqa and tested on \vqavii, and models fine-tuned on \gqa and tested on \gqa, human evaluation yields lower accuracy than automatic evaluation! This is not as expected. Upon examining some examples for responses with $100.0$ automatic accuracy but marked as \emph{incorrect} by at least 4 human raters, we again find some noise in the ground-truth answers. \cref{table:false_positives} shows some examples. Below we report the number of questions where at least 4 human raters voted incorrect even though the automatic metric indicated >=90.0 accuracy. Generative case: \{\gqa $\rightarrow$ \gqa (fine-tuned on \gqa, tested on \gqa): 86, \gqa  $\rightarrow$ \vqavii: 49, \vqavii  $\rightarrow$ \vqavii: 48\}, Discriminative case: \{\gqa $\rightarrow$ \gqa: 128, \gqa $\rightarrow$ \vqavii: 52, \vqavii $\rightarrow$ \vqavii: 76\}.

\begin{center} 
\begin{table*} 
\begin{tabular}{|p{0.18\textwidth}|p{0.4\textwidth}|p{0.355\textwidth}|}

\includegraphics[height=0.27\textwidth]{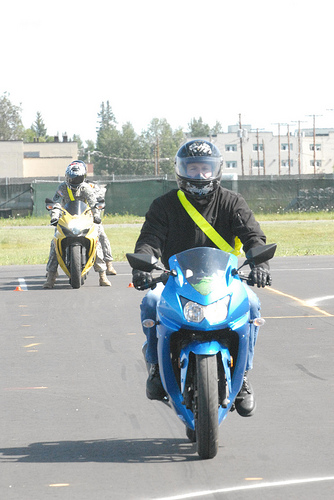} & \includegraphics[height=0.27\textwidth]{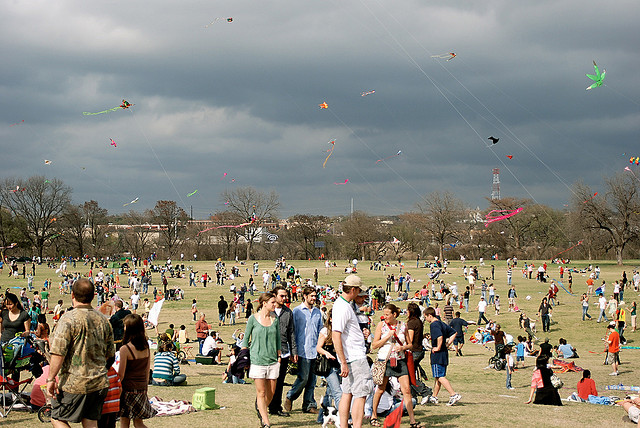} & \includegraphics[height=0.27\textwidth]{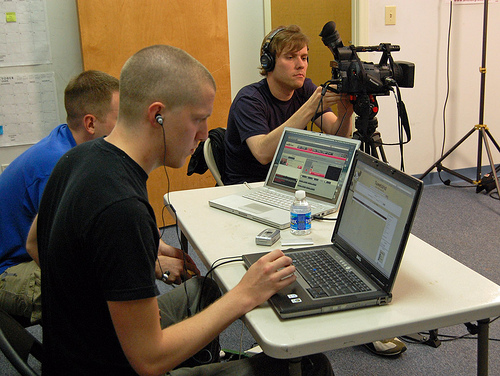} \\ 
dataset: \vg  & dataset: \vqavii & dataset: \gqa \\ 
img\_id: 2413078  & img\_id: 546983 & img\_id: 2413903 \\ 
q\_id: 151766 & q\_id: 546983002 & q\_id: 5199731 \\ 
Q: What are they wearing on their heads? & Q: What is flying in the
sky? & Q: Which kind of device is
on the table? \\ 
A: helmet  & A: kite & A: laptop \\ 
GT: helmets  & GT: kites & GT: cell phone \\ 
accurarcy: 0.0  & accurarcy: 0.0 & accurarcy: 0.0 \\ 
votes: 5 yes, 0 no  & votes: 5 yes, 0 no & votes: 5 yes, 0 no \\ 

\end{tabular} 
\caption{A few examples of questions to which the model gave a response that was objectively correct, yet the automatic evaluation metric has marked these data points as 0\% accurate. (\emph{votes} here refers to how many raters selected \emph{yes} (i.e. correct) or \emph{no} (i.e. incorrect) when asked about this data point, while GT stands for \emph{ground truth}.)} 
\label{table:5_yes} 
\end{table*} 
\end{center}

\begin{center} 
\begin{table*} 
\begin{tabular}{|p{0.28\textwidth}|p{0.20\textwidth}|p{0.45\textwidth}|}
\includegraphics[height=0.3\textwidth]{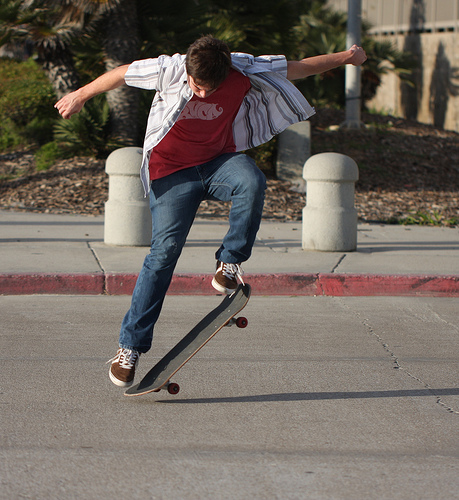} & \includegraphics[height=0.3\textwidth]{figs/COCO_val2014_000000254750.jpeg} & \includegraphics[height=0.3\textwidth]{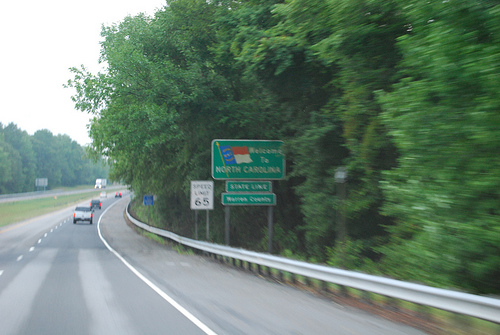} \\ 
dataset: \vg  & dataset: \vqavii & dataset: \gqa \\ 
img\_id: 2358330  & img\_id: 254750 & img\_id: 2338989 \\ 
q\_id: 700783 & q\_id: 254750003 & q\_id: 17319928 \\ 
Q: Where is he riding? & Q: Where is the toilet paper? & Q: What is on the green sign? \\ 
A: park  & A: bathroom & A: word \\ 
GT: in street  & GT: on sink & GT: flag \\ 
accuracy: 0.0  & accuracy: 0.0 & accuracy: 0.0 \\ 
votes: 3 yes, 2 no  & votes: 3 yes, 2 no & votes: 3 yes, 2 no \\ 

\end{tabular} 
\caption{Low agreement due to ambiguity. In many cases, whether an answer is correct could be up to interpretation. (\emph{votes} here refers to how many raters selected \emph{yes} (i.e. correct) or \emph{no} (i.e. incorrect) when asked about this data point, while GT stands for \emph{ground truth})} 
\label{table:ambiguous} 
\end{table*} 
\end{center}

\begin{center} 
\begin{table*} 
\begin{tabular}{|p{0.20\textwidth}|p{0.38\textwidth}|p{0.36\textwidth}|}
\includegraphics[height=0.27\textwidth]{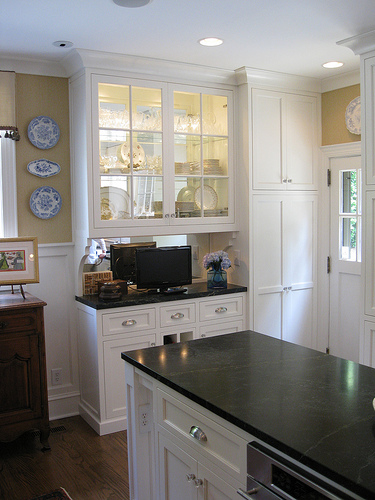} & \includegraphics[height=0.27\textwidth]{figs/COCO_val2014_000000503518.jpeg} & \includegraphics[height=0.27\textwidth]{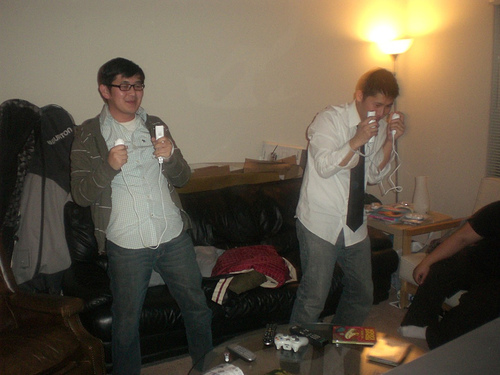} \\ 
dataset: \vg  & dataset: \vqavii & dataset: \gqa \\ 
img\_id: 2396675  & img\_id: 503518 & img\_id: 2346071 \\ 
q\_id: 1453804 & q\_id: 503518006 & q\_id: 5863992 \\ 
Q: What is the kitchen dresser? & Q: What is happening? & Q: What kind of furniture is playing a game? \\ 
A: cabinet  & A: phone & A: table \\ 
GT: brown  & GT: watching videos, showing phone & GT: couch \\ 
accuracy: 0.0  & accuracy: 0.0 & accuracy: 0.0 \\ 
votes: 2 yes, 3 no  & votes: 2 yes, 3 no & votes: 2 yes, 3 no \\ 

\end{tabular} 
\caption{Low agreement due poor question quality. Some questions have poor phrasing that make it difficult to understand what exactly is being asked. In these cases even the humans are not sure what to answer. (\emph{votes} here refers to how many raters selected \emph{yes} (i.e. correct) or \emph{no} (i.e. incorrect) when asked about this data point, while GT stands for \emph{ground truth})} 
\label{table:poor} 
\end{table*} 
\end{center} 

\begin{center} 
\begin{table*} 
\begin{tabular}{|p{0.215\textwidth}|p{0.485\textwidth}|p{0.24\textwidth}|}

\includegraphics[height=0.32\textwidth]{figs/COCO_val2014_000000367228.jpeg} & \includegraphics[height=0.32\textwidth]{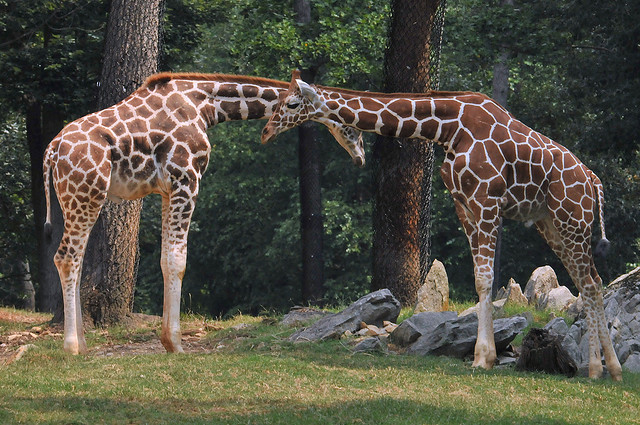} & \includegraphics[height=0.32\textwidth]{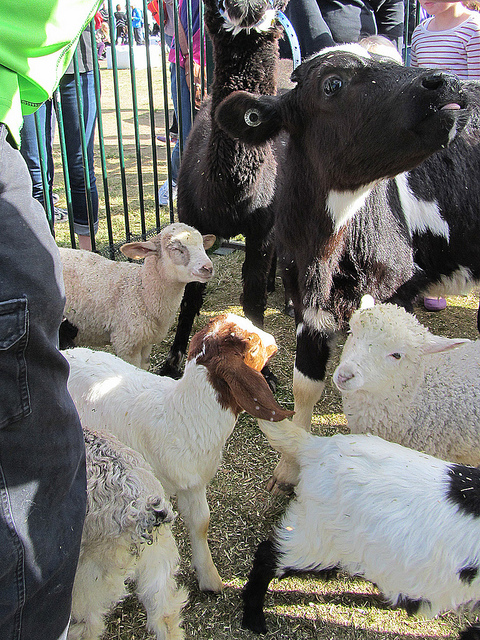} \\ 
dataset: \vqavii  & dataset: \vqavii & dataset: \vqavii \\ 
img\_id: 367228  & img\_id: 197745 & img\_id: 264737 \\ 
q\_id: 367228001 & q\_id: 197745007 & q\_id: 264737002 \\ 
Q: Is the kite flying high enough? & Q: How many spots are on this animal? & Q: How many animals are in the picture? \\ 
A: yes  & A: 100 & A: 6 \\ 
GT: [no, yes, yes, no, no, no, yes, no, no, yes]  & GT: [70, 100, 100, numerous, 200, 100, 100, 100, 20, lots] & GT: [7, 6, 6, 9, 6, 6, 6, 7, 7, 6] \\ 
accuracy: 100.0  & accuracy: 100.0 & accuracy: 100.0 \\ 
votes: 1 yes, 4 no  & votes: 1 yes, 4 no & votes: 1 yes, 4 no \\

\end{tabular} 
\caption{Examples of the few cases where humans considered the response incorrect despite 100.0 automatic accuracy. (\emph{votes} here refers to how many raters selected \emph{yes} (i.e. correct) or \emph{no} (i.e. incorrect) when asked about this data point, while GT stands for \emph{ground truth})} 
\label{table:false_positives} 
\end{table*} 
\end{center}

\end{document}